\documentclass[final,3p,times]{elsarticle}
\usepackage{amssymb}
\usepackage{amsmath}
\usepackage{booktabs}
\usepackage{rotating}
\usepackage{graphicx}
\usepackage{multirow}
\usepackage{multicol}
\usepackage{algorithm}
\usepackage{algpseudocode}
\usepackage{tikz}
\usepackage{adjustbox}
\usepackage{enumitem}
\usepackage{array}
\usepackage{url}
\usepackage{overpic}

\journal{}
\allowdisplaybreaks

\begin{document}

\begin{frontmatter}



\title{Explainable Hierarchical Deep Learning Neural Networks (Ex-HiDeNN)}

\author[label1]{Reza T. Batley}
\author[label2]{Chanwook Park}
\author[label2,label3]{Wing Kam Liu}
\author[label1]{Sourav Saha}
\affiliation[label1]{organization={Kevin T. Crofton Department of Aerospace and Ocean Engineering, Virginia Polytechnic and State University},
            addressline={1600 Innovation Drive}, 
            city={Blacksburg},
            postcode={24060}, 
            state={VA},
            country={United States}}

\affiliation[label2]{organization={Department of Mechanical Engineering, Northwestern University},
            addressline={Sheridan Road}, 
            city={Evanston},
            postcode={60208},
            state={IL},
            country={United States}}       

\affiliation[label3]{organization={Co-Founder of HIDENN-AI, LLC},
            addressline={1801 Maple Ave}, 
            city={Evanston},
            postcode={60201}, 
            state={IL},
            country={United States}} 

\begin{abstract}
Data-driven science and computation have advanced immensely to construct complex functional relationships using trainable parameters. However, efficiently discovering interpretable and accurate closed-form expressions from complex dataset remains a challenge. The article presents a novel approach called Explainable Hierarchical Deep Learning Neural Networks or Ex-HiDeNN that uses an accurate, frugal, fast, separable, and scalable neural architecture with symbolic regression to discover closed-form expressions from limited observation. The article presents the two-step Ex-HiDeNN algorithm with a separability checker embedded in it. The accuracy and efficiency of Ex-HiDeNN are tested on several benchmark problems, including discerning a dynamical system from data, and the outcomes are reported. Ex-HiDeNN generally shows outstanding approximation capability in these benchmarks, producing orders of magnitude smaller errors compared to reference data and traditional symbolic regression. Later, Ex-HiDeNN is applied to three engineering applications: a) discovering a closed-form fatigue equation, b) identification of hardness from micro-indentation test data, and c) discovering the expression for the yield surface with data. In every case, Ex-HiDeNN outperformed the reference methods used in the literature. The proposed method is built upon the foundation and published works of the authors on Hierarchical Deep Learning Neural Network (HiDeNN) and Convolutional HiDeNN. The article also provides a clear idea about the current limitations and future extensions of Ex-HiDeNN.         
\end{abstract}


\end{frontmatter}
\section{Introduction}
\label{sec1}
Complexity is an inescapable facet of real-world engineering systems. Input-output relationships in such systems are rarely linear or amenable to simple analytical forms \cite{Ljung1999}. This inherent complexity often defies accurate representation by methods such as ordinary least-squares regression (OLS) \cite{Legendre1805}, \cite{Gauss1821}, and general additive models (GAMs) \cite{Hastie1986}. To overcome this expressive gap, modern neural architectures - multilayer perceptrons (MLPs) \cite{Rumelhart1986}, convolutional neural networks (CNNs) \cite{Lecun1998}, graph neural networks (GNNs) \cite{Scarselli2009} and beyond – have risen to prominence. These models stack many layers of nonlinear transformations and vast numbers of trainable parameters to capture intricate relationships \cite{Lou2023}. Artificial neural networks have successfully been used in, amongst many other things, aerodynamic modeling with MLPs \cite{Espinosa2023}, reduced-order flow field reconstructions with CNNs \cite{Eichinger2022}, and rapid mesh-based physics simulations with GNNs \cite{Pfaff2020}. Despite their impressive predictive power, their tendency towards opacity - formally as ``black-box models" - has spurred research into alternatives. In machine learning, a black-box model is one whose internal workings are not easily accessible or interpretable \cite{Hassija2024}. For this reason, recent developments such as Kolmogorov-Arnold Networks (KANs) \cite{liu2025kan}, interpolating neural networks (INNs) \cite{Park2024, Saha2024}, and KHRONOS \cite{batley2025khronos} have sought to build architectures with underlying mathematical structure, offering a degree of baked-in interpretability. Nevertheless, these models remain translucent, offering only partial insight into the relationships they learn. This class of models is known as ``grey-box”.

Despite this increasing degree of interpretability, grey- and black-box models are discouraged or even deemed outright unsuitable in many safety-critical applications \cite{Rudin2019}. For example, NASA’s AI assurance guidance highlights that the lack of transparency in ML-enabled systems impedes the rigorous verification and validation process \cite{NASA}. The Federal Aviation Administration (FAA)’s safety assurance techniques assume that a designer can explain every aspect of their system’s design, but such explanations are not readily extendable to black box AI models, creating a challenge in assuring their safe operation \cite{FAA}. These sentiments are echoed in the joint FDA-Health Canada-Medicines and Healthcare Products Regulatory Agency (MHRA) guidance on AI \cite{FDA}. There is clearly a greater need for interpretable AI models. Interpretability is the degree to which a human can understand the cause of a decision, an interpretable method being one with which a user can correctly and efficiently predict its results \cite{Biran2017, Miller2019}.

In pursuit of interpretable machine learning, we - going full circle - turn to symbolic surrogates \cite{Marcin2023}. These are simply closed-form formulas typically composed of elementary functions, completely transparent to the user. Explicit expressions allow for direct comparison with known physical mechanisms and the identification of dominant effects and variables. The use of symbolic surrogates – learning expressions directly from data – is not new, in fact tracing back to the 1990s, first performed with genetic algorithms \cite{Koza1994}. More modern approaches \cite{Schmidt2010, Brunton2016} can also be used to learn governing equations for dynamical systems with remarkable accuracy. Beyond explicit interpretability, closed-form expressions offer several additional benefits:
\begin{itemize}
\item
Parsimony through training: expression complexity can be penalized during training to yield more compact models that tend to generalize better \cite{Brunton2016}. 
\item
Parsimony through pruning: expressions can be easily pruned as part of post-processing, unlocking similar benefits \cite{Schmidt2010}.
\item
Physical consistency: invariances, symmetries and homogeneity, can be verified or even enforced \cite{Udrescu2020}.
\item
Seamless integration: symbolic surrogates plug directly into existing simulation, optimization and control pipelines \cite{Bahmani2024}, as well as embedment into multiscale systems. 
\item
Closed-form formal analysis: error bounds, bifurcation studies, and asymptotic analysis are a few amongst a plethora of techniques that can be used with full mathematical control. 
\end{itemize}

Despite these strengths, current symbolic regression approaches face critical challenges. First and foremost, finding an optimal symbolic form is an NP-hard problem, with the search space growing exponentially in the number of variables allowed operators. In other words, even when made tractable, the combinatorial complexity of symbolic regression is exponential. Further, as is laid out in \cite{Petersen2019, Bahmani2024}, models lack consistency and can routinely fail to recover simple expressions, holding especially true when problem dimensionality increases. Additionally, symbolic regression can overfit to noise – especially the case with genetic programming \cite{Raghav2024}. For efficient, high-dimensional symbolic regression, we require a framework that can minimize combinatorial blow-up of the search space, filter noise, and provide reproducible results. 

\begin{figure}
    \centering
    \includegraphics[width=\linewidth]{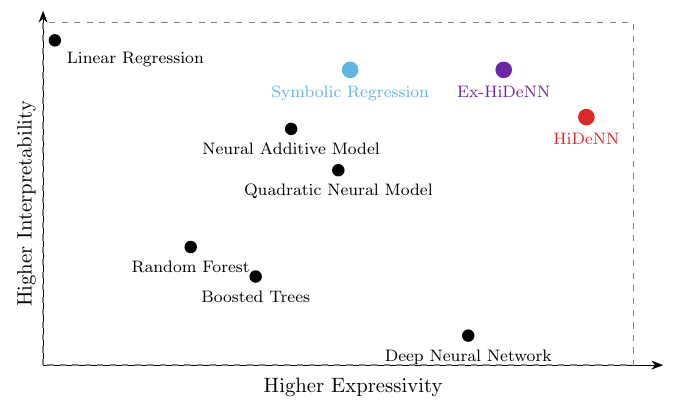}
    \caption{A rough guide to expressivity vs interpretability of common machine learning approaches, adapted from \cite{Bahmani2024}}
    \label{fig:fig1}
\end{figure}

Introducing Ex-HiDeNN: Explainable Hierarchical Deep Learning Neural Networks. This novel, hybrid architecture combines the expressivity and separability of C-HiDeNN-TD or equivalently, an interpolating neural network \cite{Park2024, Saha2024}, with the inherent interpretability of symbolic regression; automatically and efficiently extracting closed-form expressions from data. This approach mitigates combinatorial blow-up, smooths noise and ensures consistent recovery. Ex-HiDeNN's position in the current expressivity vs interpretability paradigm is illustrated in figure \ref{fig:fig1}. It is crucial to define the terms ``expressivity" and ``interpretability" in this work.

\textbf{Expressivity} refers to the demonstrated effectiveness of a model in learning an accurate and generalizable representation of a target function from given data. This practical definition inherently values models that do not ``squander" their capacity by fitting to noise, a common occurrence for traditional black-box models.
Under this definition, deep neural networks, despite being universal approximators, often exhibit lower expressivity on scientific problems. The enormous functional space it learns in makes the learning process inefficient and prone to finding local minima or overfitting on sparse data. Conversely, while HiDeNN cannot represent a function space quite as large as a deep neural network, its architecture is endowed with a smaller functional space derived from principles of computational mechanics. It therefore tends to demonstrate higher expressivity, especially in scientific problems.

\textbf{Interpretability} is the degree to which a human can understand a model's decision-making process and predict its results. This work considers parsimonious symbolic expressions as the highest form of interpretability, with deep neural networks the lowest. Deep neural networks simply lack an understandable, outward ``logic" in their decision-making processes.

The detailed algorithm is laid out in Section \ref{sec2}, but at a high level, Ex-HiDeNN automates a two-stage pipeline guided by a separability score assigned to the data. The first stage is the hierarchical deep learning neural network (C-HiDeNN-TD), a novel architecture that uses tensor decomposition for separability. The C-HiDeNN-TD learns a unidimensional representation for each feature. These combine in an outer product to form a single multi-dimensional component, called a mode. Several of such modes are initialized, trained, and added to form the global representation. The global representation is continuously differentiable, allowing us to manipulate its derivatives, used to quantify local separability. The C-HiDeNN-TD structure uses a hierarchical neural architecture proposed in HiDeNN \cite{Saha2021}, a convolutional patch function \cite{Park2023}, and tensor decomposition \cite{lu2023convolution,li2023convolution,zhang2022HiDeNN,zhang2021hierarchical}. Ex-HiDeNN selects a sampling strategy based on the separability score. For high-scoring data, the surrogate is sampled per-dimension, each fit with univariate expression. These expressions are then multiplied to form the final expression. For medium scores, sampling is done on a per-dimension basis per-mode, constructing an additive set of multiplicative terms. For low scores, the interpolant is sampled globally. Symbolic regression then proceeds as standard. 

The remainder of this paper is organized as follows. Section \ref{sec2} introduces the three core components of Ex-HiDeNN: C-HiDeNN-TD's architecture \ref{sec:C-HiDeNN-TD}, PySR \ref{sec:pysr}, and the separability measure \ref{sec:hessian}, as well as the overall pipeline \ref{sec:Ex-HiDeNN}. Section \ref{sec3} presents numerical experiments on synthetic benchmarks. A simple demonstration in \ref{sec:demo} and evaluation on several benchmark functions \cite{Vlad2009} - compared to a contemporary genetic approach, Symbolic KAN \cite{liu2024kan20} and PySR by itself - in \ref{sec:bench}. Additionally, Ex-HiDeNN's ability to discover a known multi-dimensional dynamical system is demonstrated in \ref{sec:dyn}. Ex-HiDeNN is then applied to current engineering problems with unknown physics in Section \ref{sec4}. In \ref{sec:fatigue} Ex-HiDeNN discovers a closed-form equation for fatigue from data in additively manufactured steel. In \ref{sec:hardness} Ex-HiDeNN discovers an equation for hardness from microindentation data. In \ref{sec:yield} Ex-HiDeNN learns a yield surface expression for a pressure sensitive Matsuoka-Nakai material. Sections \ref{sec5}, \ref{sec6} conclude with a discussion of limitations and avenues for future work.

In summary, this work makes four key contributions,
\begin{itemize}
    \item introduction of a pointwise Hessian-based separability score $S_\otimes$ computed via automatic differentiation of the C-HiDeNN-TD-derived surrogate;
    \item introduction of a novel two-stage AI framework that marries a continuously differentiable - and structured - surrogate (C-HiDeNN-TD) with symbolic regression (PySR) to extract closed-form expressions from data;
    \item comprehensive validation: Ex-HiDeNN is validated on synthetic benchmarks and real engineering datasets to demonstrate its comparable or superior accuracy, interpretability and efficiency;
    \item discovery of novel closed-form models for i) fatigue life in additively manufactured steel, ii) material hardness from microindentation data, and iii) the Matsuoka-Nakai yield surface directly from data.
\end{itemize}
\section{Methodology}
\label{sec2}
In what follows, model performance is frequently evaluated using the mean-square error (MSE), root-mean-square error (RMSE), maximum absolute error (MaxAE), relative error (RE) and coefficient of determination, or $R^2$(-score). Furthermore, all data used comes in the form of input-output pairs, $(\boldsymbol{x}_i,u_i)$, $i=1,\dots,N$, $\boldsymbol{x}=(x_1,\dots,x_n)^T\in \mathbb{R}^n,u\in\mathbb{R}$. Let $\hat u(\boldsymbol{x}_i)$ be the prediction of a model on the $i$-th input $\boldsymbol{x}_i$. Then, 
\begin{align}
    \textrm{MSE}&=\frac1N\sum_{i=1}^N(\hat u(\boldsymbol{x}_i)-u_i)^2, \\
    \textrm{RMSE}&=\sqrt{\textrm{MSE}}, \\
    \textrm{MaxAE}&=\max\limits_{i=1,\dots,N}|\hat u(\boldsymbol{x}_i)-u_i|\\
    \textrm{RE}&=\frac1N\sum_{i=1}^N\frac{|\hat u(\boldsymbol{x}_i)-u_i|}{|u_i|}, \\
    R^2&=1 - \frac{\sum\limits_{i=1}^N (u_i - \hat u(\boldsymbol{x}_i))^2}{\sum\limits_{i=1}^N (u_i - \bar u)^2},~\bar u=\sum\limits_{i=1}^Nu_i.
\end{align}
When sampling the surrogate, a combination of Latin Hypercube Sampling (LHS) \cite{mckay1979lhs} and local perturbation sampling (LPS) is typically used. LHS involves dividing the input domain into equal-probability intervals along each axis and drawing one point uniformly at random from each interval. LPS is a form of data augmentation sampling, achieved by generating local perturbations in the neighborhood of training samples. This combination of techniques, while heuristic, gives both global, stratified coverage from LHS and local accuracy from LPS.

\subsection{C-HiDeNN-TD}
\label{sec:C-HiDeNN-TD}
This section introduces the neural network surrogate \cite{Park2024, Saha2024, mostakim2025inntl} part of Ex-HiDeNN, responsible for learning a continuously differentiable surrogate that interpolates the target domain. This neural network is called C-HiDeNN-TD, and offers scalability via tensor decomposition in input parameters and adaptable, nonlinear interpolation techniques for high accuracy in downstream applications. In particular, such scalability is crucial in symbolic regression, an NP-hard problem. In suitable problems, C-HiDeNN-TD generates a decomposable surrogate and samples each of the one-dimensional decompositions. Symbolic regression can then be performed independently on each sample, drastically reducing the effective search space and time complexity, while also improving parallelism.

\begin{figure}
\centering
\includegraphics[width=0.85\linewidth]{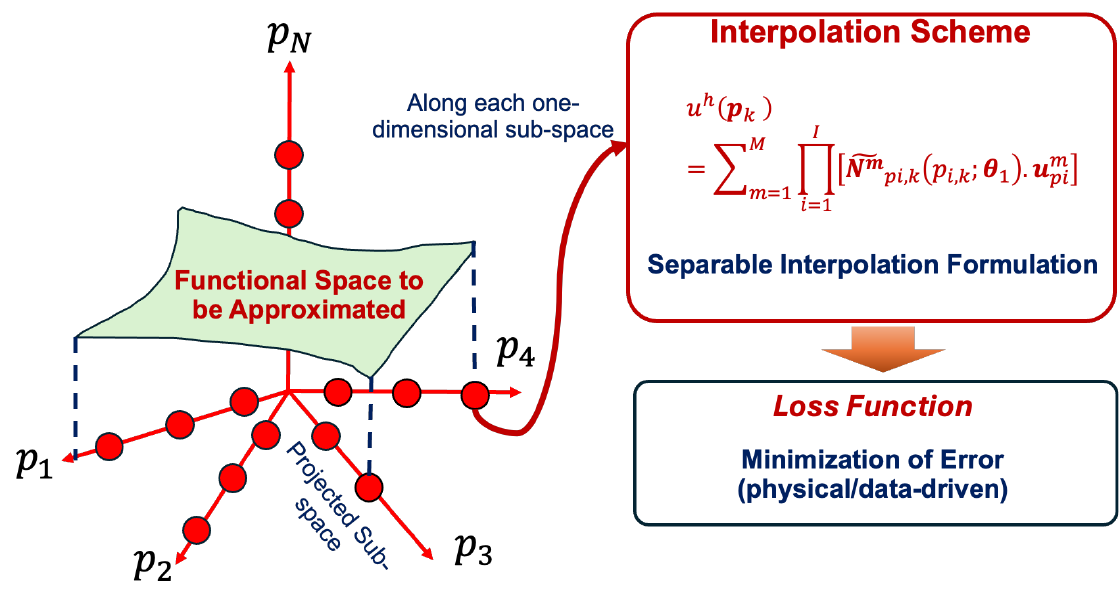}
\caption{A schematic diagram describing the separable neural network architecture of C-HiDeNN-TD.}
\label{fig:fig2}
\end{figure}

The idea of the proposed network architecture is shown in Figure \ref{fig:fig2}. Assume a function $u(.)$ that has $I$ inputs $[p_1,p_2,\ldots,p_I]$, and each input domain is sampled with $k$ points or nodes, the mapping equation can be written as

\begin{equation}\label{eqn:6}
u^h(\mathbf{p}_k)=u^h(p_{1,k},p_{2,k},...,p_{I,k})= \sum_{k=1}^{K}\prod_{i=1}^I\sum_{m=1}^{M}\tilde N_{pi,k}(p_{i,k};\mathbf{\theta})\cdot u_{pi,k}^m.
\end{equation}

Here, $\mathbf{p}_{k}=[p_{1,k}, p_{2,k}, ..., p_{I,k}]$ is the set of input variables of $k^{th}$ training data, and $u^h(\cdot)$ is the learned function. The index $i$ represents parameter dimensions as $1,2,...,I$. The index $m$ indicates the number of modes. These modes can be arbitrarily fixed, and act as \textit{hyperparameters}. The central idea is to split the functional space into several one-dimensional sub-spaces. Along each such one-dimensional subspace, a highly pruned neural network interpolant ${\tilde N}_{pi,k}(p_{i,k};\mathbf{\theta})$ - akin to a meshfree formulation - is constructed as shown in Figure \ref{fig:fig2}. C-HiDeNN-TD has trainable parameters embedded in its structure \cite{Saha2021, Park2023}. The interpolants are based on pre-selected interpolation nodes. Additionally, each input sub-space is discretized into multiple segments. The interpolation nodes are assigned at the boundaries of these segments just like finite element method. The interpolant along any parameter $p$ and any segment $s$ of $p$ is written as,

\begin{equation}
\tilde{N}_p(\cdot) = \sum_{k\in A^s} N_{p,k}(.)\sum_{l\in A^r}W_{r,a,pd,l}^k(.).    
\end{equation}

In this equation, the first part of the interpolant is similar to a finite element shape function and can be linear or non-linear as the case requires. The next part of the shape function is an enrichment following radial point method in meshfree literature. The former shape function acts on $k$ nodes inside each segment ($A^s$). The enriched part $W_{r,a,pd,l}^l(.)$, which is called the \textit{patch function}, aims to incorporate information from neighboring nodes within a neighborhood near a segment node. The patch size $r$ determines this neighborhood. Other network parameters are the polynomial order $pd$ of the approximation function inside the patch and dilation parameter $a$. These patch and interpolation functions can be constructed in any arbitrary manner as hyperparameters. All the trainable parameters are grouped as $\mathbf{\theta}$ in Equation \ref{eqn:6}. For more details, the readers are referred to \cite{Park2024, Saha2024}. Finally, the loss function of C-HiDeNN-TD can be written as

\begin{equation}
\displaystyle{\min_{\mathbf{\mathbf{u}_k, \boldsymbol{\theta}}} \cal L}= \frac{1}{N}\sum_{n=1}^{N}(\sum_{k=1}^{K}\prod_{i=1}^{I}\sum_{m=1}^{M}[\tilde N_{pi,k}(p_{i,k};\boldsymbol{\theta}) \cdot \mathbf{u}^{m}_{pi,k}] - u_n^{train})^2.
\end{equation}

Here, $u_n^{train}$ are the training values at $N$ data points. An important note is that data points are not ``nodes", rather the given samples for each input variable.

\subsection{Hessian Separability Measure}
\label{sec:hessian}
Analysis of the Hessian matrix of a multivariate function can determine whether it is sufficiently multiplicative to be approximated by a tensor decomposition. In this section, an $n$-dimensional scalar field $f:\mathbb{R}^n\rightarrow\mathbb{R}$ is defined so that
\begin{align}
    \label{e:f}
    f = f(\boldsymbol{x}),~\boldsymbol{x}=(x_1,...,x_n)^T\in\mathbb{R}^n.
\end{align}

\paragraph{The Hessian Matrix}
The Hessian matrix is a square matrix of a function's second partial derivatives which describes the function's local curvature. Componentwise, the Hessian matrix of $f$ is defined 
\begin{align}
    H_{ij}=\frac{\partial^2 f}{\partial x_i\partial x_j}.
\end{align}
As a full matrix, this looks like
\begin{align}
    H_f(\boldsymbol{x})=
    \left(
    \begin{array}{cccc}
        \frac{\partial^2 f}{\partial x_1\partial x_1} & \frac{\partial^2 f}{\partial x_1\partial x_2}  & \cdots & \frac{\partial^2 f}{\partial x_1\partial x_n}\\
        \frac{\partial^2 f}{\partial x_2\partial x_1} & \frac{\partial^2 f}{\partial x_2\partial x_2} &  \cdots & \frac{\partial^2 f}{\partial x_2\partial x_n} \\
        \vdots & \vdots & \ddots & \vdots \\
        \frac{\partial^2 f}{\partial x_n\partial x_1} & \frac{\partial^2 f}{\partial x_n\partial x_2} & \cdots & \frac{\partial^2 f}{\partial x_n\partial x_n}
    \end{array}
    \right)
\end{align}

\paragraph{Additive Separability}
A function $f$ as in \eqref{e:f} is additively separable if it admits a decomposition of univariate functions of each of its variables,
\begin{align}
    f(x_1,\dots,x_n)=f_1(x_1)+\dots+f_n(x_n).
\end{align}
It follows that
\begin{align}
    \frac{\partial^2 f}{\partial x_i\partial x_j} = \delta_{ij}f_i''(x_i),
\end{align}
where $\delta_{i,j}=1$ if $i=j$ else $0$ is the Kronecker delta. Its Hessian matrix is therefore
\begin{align}
    H_f(\boldsymbol{x})=
    \left(
    \begin{array}{cccc}
        f_1''(x_1) & 0 & \cdots & 0 \\
        0 & f_2''(x_2) & \cdots & 0 \\
        \vdots & \vdots & \ddots & \vdots \\
        0 & 0 & \cdots & f_n''(x_n)
    \end{array}
    \right),
\end{align}
so for an additively separable $f$, all off-diagonal terms vanish. The diagonal dominance of a function's Hessian can therefore give us insight into how additively separable that function is. 
\paragraph{Multiplicative Separability}
A function $f$ is multiplicatively separable if it can be written as a product of univariate functions of each of its variables:
\begin{align}
    f(x_1,\dots,x_n)=f_1(x_1)f_2(x_2)\dots f_n(x_n).
\end{align}
Application of a log-modulus transformation gives us
\begin{align}
    \log|f(x_1,\dots x_n)|=\log|f_1(x_1)|+\log|f_2(x_2)|+\dots+\log|f_n(x_n)|,
\end{align}
an additively separable function. Thus, analysis of $H_{\log|f|}(\boldsymbol{x})$ can give qualitative insight into the \textit{multiplicative} separability of $f$.

\subsubsection{Numerical Separability Quantification}

Ex-HiDeNN does not deal with analytical inputs, rather numerical data in the form of input-output pairs, $(\boldsymbol{x}_i, y_i)$. Therefore, the above approach has to be adapted to this end. This can be done as part of preprocessing by fitting the data to a grid, calculating a numerical Hessian, and averaging it down to block form. Since Ex-HiDeNN already learns an automatically differentiable interpolant, $\hat f_\text{C-HiDeNN-TD}$, this can be used to calculate an extremely accurate Hessian. 

The multiplicativity score, $S_\otimes$, is then calculated by measuring the square-relative contribution of diagonal terms. For a Hessian $H_{\log|f|}$ with diagonal $D_{\log|f|}=\textrm{diag}(H_{\log|f|})$, this can be written in terms of the Frobenius norm, $\| \cdot \|_F$, so that
\begin{align}
    S_\otimes = \frac{\|D_{\log|f|}\|^2_F}{\|H_{\log|f|}\|^2_F}.
\end{align}

\begin{figure}
    \centering
\includegraphics[width=0.6\linewidth]{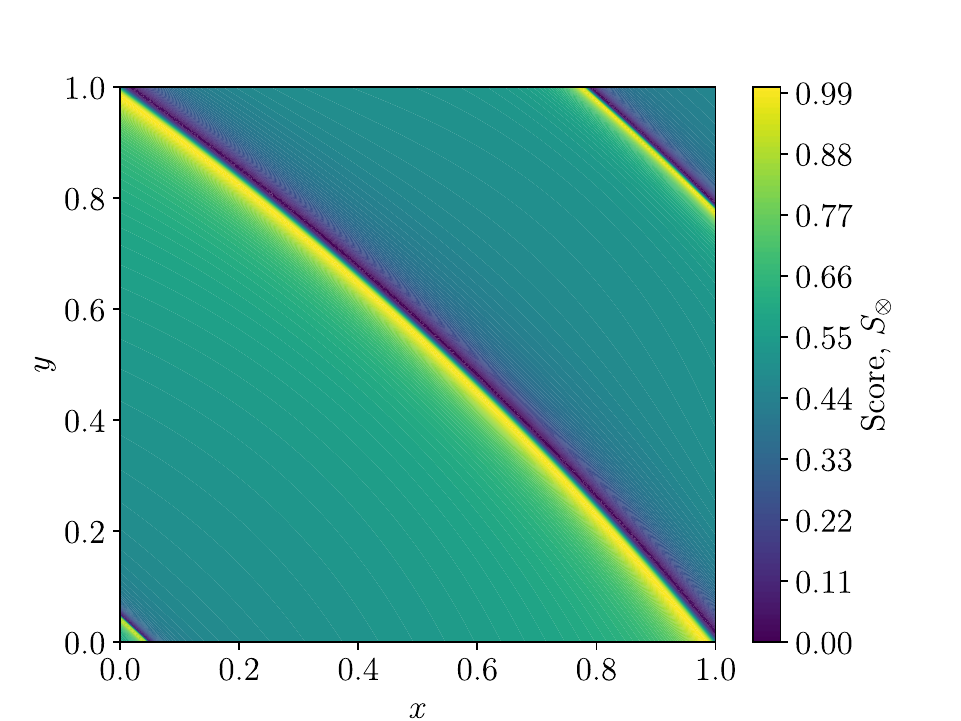}
    \caption{Local multiplicative separability $S_\otimes(x,y)$ for the test function \eqref{e:egsep} on $[0,1]^2$. High values (yellow) indicate near-perfect separability, while low values (purple) show strong coupling. The domain mean $\bar S_\otimes(x,y)=0.52.$}
    \label{f:fig3}
\end{figure}

The score $S_\otimes$ is bounded within $[0, 1]$, with $S_\otimes=1$ when $f$ or $\hat f_\text{C-HiDeNN-TD}$ are perfectly multiplicatively separable, and $S_\otimes=0$ when they are purely coupled. This logic is applied pointwise, for local separability scores, but can be extended to a domain by taking a mean over samples. An example of a map of these local separability scores can be seen in \ref{f:fig3}, taken from a function
\begin{align}
    \label{e:egsep}
    f(x,y)=(x-3)(y-3)+2\sin((x-4)(y-4))
\end{align}
defined on the unit square $[0,1]^2$. The pointwise score ranges from 0 to 1, and the mean separability score $\bar S_\otimes(x,y)$ is 0.52, indicating low-to-moderate separability over the entire domain.

\subsection{Symbolic Regression (PySR)}
\label{sec:pysr}
Once the C-HiDeNN-TD surrogate is trained and a mean separability score $S_\otimes$ computed, Ex-HiDeNN invokes a symbolic regression engine to recover closed-form expressions in the assigned sampling regime. In this primordial implementation, it uses PySR \cite{PySR}: an open source, multi-population evolutionary algorithm built on \texttt{SymbolicRegression.jl}, a high-performance Julia library. PySR is mostly built as a simple evolutionary algorithm as follows:
\begin{enumerate}
    \item Initialize a population of $N$ individuals, $\{I_1, I_2, \dots, I_N\}$, a fitness function, $f$, and a set of mutation operators.
    \item Randomly select a $k$-large subset of individuals from the population. Classically, one would take $k=2$.
    \item Run a tournament by evaluating the fitness of each individual, with respect to the fitness function, in that subset. This sorts the subset from fittest to least fit.
    \item Consider the fittest candidate. With probability $p$, accept it as a "parent", discarding it with probability $1-p$. If discarded, move on to the second fittest candidate and repeat the process. If the list is exhausted without acceptance, take the final individual. 
    \item Copy the selected individual, and apply a randomly mutation operation from the pre-defined set.
    \item Replace a member of the population, typically the weakest, with this mutated individual.
\end{enumerate}
PySR departs from this classical loop via two main modifications. Firstly, \textit{age-regularized} evolution: replacing the eldest member of the population, rather than the least fit, in step 6). This change prevents the population from specializing too quickly and getting stuck in a local minimum. Next, PySR applies \textit{simulated annealing} to step 4). This rejects mutation with probability $p$ depending on the potential change in fitness. Specifically, 
\begin{align}
    p = \exp\left(\frac{L_F-L_E}{\alpha T}\right),
\end{align}
$L_E$ and $L_F$ being the fitness pre- and post- the potential mutation, $\alpha$ and $T\in[0, 1]$ being hyperparameters. Note that $p$ is the same probability that controls population diversity in step 4).
Unless specified otherwise - and with the key exceptions of maximum complexity, population size and maximum iterations - all PySR hyperparameters are left as the out-of-the-box defaults. For a complete list of hyperparameters and their default values, readers are encouraged to visit the PySR documentation at \url{https://astroautomata.com/PySR/}.

\begin{algorithm}[h]
\caption{The two-stage Ex-HiDeNN pipeline}
\label{a:Ex-HiDeNN}
\begin{algorithmic}
  \Require Training data $(x_i,y_i)$, threshold $T_{S_{\otimes}}^\text{high}$, $T_{S_{\otimes}}^\text{med}$
  \Ensure Learned symbolic expression $\hat f(x)$
  \State Train C-HiDeNN-TD on $\{x_i,y_i\}$ to get $\mathrm{\hat f_\text{C-HiDeNN-TD}}$
  \State Calculate log-Hessian $H_{i,j}\gets\frac{\partial^2}{\partial x_i\partial x_j}(\mathrm{\hat f_\text{C-HiDeNN-TD}})$
  \State Calculate score $S_{\otimes}\gets\frac{\|\text{diag}(H)\|_F^2}{\|H\|_F^2}$
  \If{$S_{\otimes}\ge T_{S_{\otimes}}^\text{high}$}
    \For{$d=1,\dots,D$}
      \State $u_d\gets\text{sample}(\hat f_\text{C-HiDeNN-TD}^d)$
      \State $e_d\gets\text{PySR}(u_d)$
    \EndFor
    \State $\hat f(x)\gets \prod_{d=1}^D e_d(x_d)$
  \ElsIf{$S_{\otimes}\ge T_{S_{\otimes}}^\text{med}$}
    \For{$m=1,\dots,M$}
      \For{$d=1,\dots,D$}
        \State $u_{m,d}\gets\text{sample}(\hat f_\text{C-HiDeNN-TD}^{m,d})$
        \State $e_{m,d}\gets\text{PySR}(u_{m,d})$
      \EndFor
    \EndFor
    \State $\hat f(x)\gets \sum_{m=1}^M\prod_{d=1}^D e_{m,d}(x_d)$
  \Else
    \State $\mathcal U\gets \text{sample}(\mathrm{\hat f_\text{C-HiDeNN-TD}})$
    \State $\hat f\gets\text{PySR}(\mathcal U)$
  \EndIf
  \State \Return $\hat f(x)$
\end{algorithmic}
\end{algorithm}
\subsection{Adaptive Mathematical forms for Varying Separability}
\label{sec:Ex-HiDeNN}
Ex-HiDeNN adapts its final mathematical structure to best match the inherent nature of given data, in a way guided by the separability score. The framework is therefore able to propose highly parsimonious models for suitably structured data, or more expressive models for complex, interacting systems. The primary functional forms are as follows,
\paragraph{High Separability, $0.95 \lesssim S_\otimes$}
In this case of near-perfectly separable data, Ex-HiDeNN proposes a single-mode model. The final expression $\hat f(x)$ takes the form of a product of univariate functions $e_d(x_d)$,
\begin{align}
    \hat f(x)=\prod_{d=1}^D e_d(x_d).
\end{align}
Here, each $e_d$ is discovered separately, reducing an exponential search-space blow up to a linear one.
\paragraph{Moderate Separability, $0.6 \lesssim S_\otimes \lesssim 0.95$}
For moderately coupled data, Ex-HiDeNN uses a sum-of-products, multi-mode form akin to a tensor decomposition. The final expression is a sum of $M$ multiplicative modes,
\begin{align}
    \hat f(x)=\sum_{m=1}^M\prod_{d=1}^D e_{m,d}(x_{d}).
\end{align}
This more expressive form better captures cross-variable interactions while exploiting underlying separable structure.
\paragraph{Low Separability, $S_\otimes \lesssim 0.6$}
With strongly coupled data, the framework does not assume separable structure, rather performing full, multi-dimensional symbolic regression. The surrogate is still taken advantage of, as it allows for a more intelligent sampling strategy. In these cases, a combination of LHS and LPS is used. 

This adaptive mathematical framework is implemented in the complete two-stage pipeline as summarized in Algorithm \ref{a:Ex-HiDeNN}.
\section{Benchmark Problems}
\label{sec3}
\subsection{Demonstrative Example}
\label{sec:demo}
This first problem demonstrates Ex-HiDeNN's ability to discover an expression for an unknown function from data. The goal is to reconstruct the function, 
\begin{align}
    \label{eq:3.1}
    f(x,y)=e^{x+2y},(x, y)^T\in\Omega,  
\end{align}

\begin{figure}[h]
    \centering
    \includegraphics[width=\textwidth]{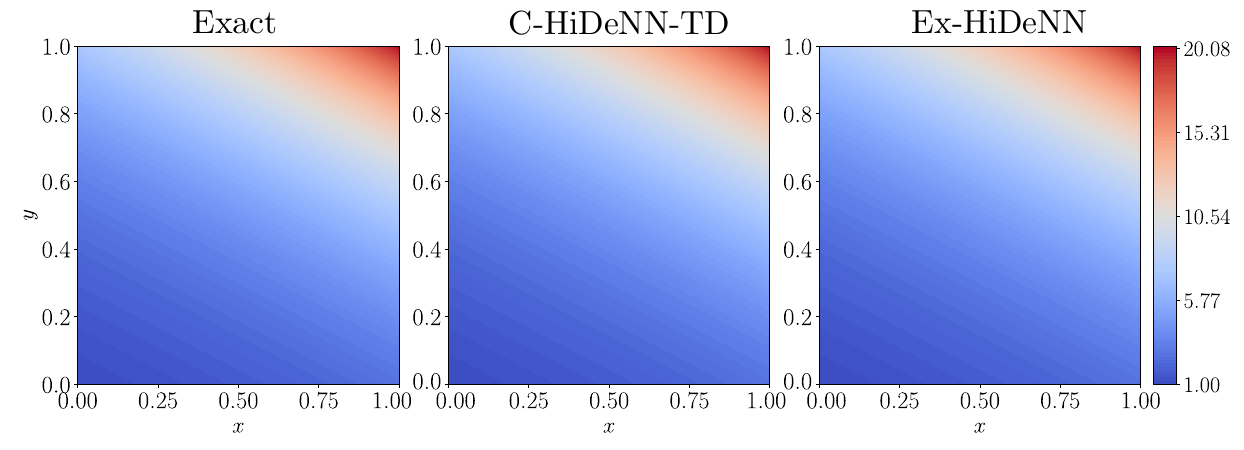}
    \caption{Demonstrative example of Ex-HiDeNN on a known, separable form \eqref{eq:3.1}.}
    \label{fig:fig4dmo}
\end{figure}

where $\Omega=[0,1]^2$. Training data is generated by sampling $f$ at 10,000 points using LHS, and then split 70/15/15 train-test-validation. This data is used to train a 42-parameter C-HiDeNN-TD is for 100 epochs. The resulting surrogate model has an RMSE of $6\times10^{-7}$. PySR then uses the data learned by the C-HiDeNN-TD surrogate in order to discover a separable formulation. It is run for 1000 iterations with a maximum complexity of 15 and 20 populations. The learned symbolic expression in the $x-$ and $y-$ directions are $e^{0.44727}e^{0.99986x}$ and $e^{-0.44719}e^{2y}$, respectively. This gives the overall expression $e^{0.0008}e^{x+2y}, $ with an RMSE of 0.00025. Figure \ref{fig:fig4dmo} shows a comparison of the plots of the exact expression (first), the C-HiDeNN-TD surrogate (second) and the symbolic reconstruction (third).

\subsection{Comparison to Contemporary Benchmarks}
\label{sec:bench}
To test our framework, a benchmark suite proposed by \citet{Vlad2009} is employed. This suite is a standard in the symbolic regression community for evaluating an algorithm's ability to recover an exact function from sampled data. The problems are designed to challenge key capabilities; how well a model handles a variety of functional forms, with the suite including a mix of polynomial, rational, and transcendental functions. Additionally, it tests how well a model extrapolates by intentionally taking testing data from a wider domain than the training set. Extrapolation is one key benefit of symbolic regression compared to surrogates designed for interpolation, and is thus important to consider. In addition, some training sets are particularly sparse. Therefore, success on these benchmarks is a strong indicator of a model's ability to find an accurate and generalizable expression from sparse data.

In particular, this set provides eight exact functions and their associated testing and training samples. We identify each function $V_1-V_8$ with their corresponding names as in \cite{Vlad2009}: Kotanchek, Salustowicz, Salustowicz2D, 5D Unwrapped Ball, Rational Polynomial 3D, SineCosine, Ripple, and Rational Polynomial 2D, respectively. Their expressions are written out in \eqref{eq:VS1}-\eqref{eq:VS8}. We denote random uniform sampling of $N\in\mathbb{N}$ points from the set $[a, b]$ as $U(a, b, N), $ and equispaced sampling from the same set with step $c>0$ $E(a, b, c)$. Note that $E(a, b, c)$ samples $N=\left \lfloor{\frac{b-a}{c}}\right \rfloor+1$ points. The sampling methods for each function are also laid out in \eqref{eq:VS1}-\eqref{eq:VS8}.

\begin{align}
    \label{eq:VS1}
    V_1:& \quad f_1(x_1,x_2) = \frac{e^{-(x_1-1)^2}}{1.2+(x_2-2.5)^2}, \notag \\
    & \quad x_{\text{train}} \sim U(0.3, 4, 100)^2, \quad x_{\text{test}} \sim E(-0.2, 4.2, 0.1)^2. \\
    V_2:& \quad f_2(x) = e^{-x}x^3\cos x\sin x(\cos x\sin^2x-1), \notag \\
    & \quad x_{\text{train}} \sim E(0.05, 10, 0.1), \quad x_{\text{test}} \sim E(-0.5, 10.5, 0.05). \\
    V_3:& \quad f_3(x_1,x_2) = (x_2-5)f_2(x_1), \notag \\
    & \quad x_{\text{train}} \sim E(0.05, 10, 0.1) \times E(0.05, 10, 2), \notag \\
    & \quad x_{\text{test}} \sim E(-0.5, 10.5, 0.05) \times E(-0.5, 10.5, 0.5). \\
    V_4:& \quad f_4(x_1,\dots,x_5) = \frac{10}{5+\sum_{i=1}^5(x_i-3)^2}, \notag \\
    & \quad x_{\text{train}} \sim U(0.05, 6.05, 1024)^5, \notag \\
    & \quad x_{\text{test}} \sim U(-0.25, 6.35, 5000) \times \{0\}^4. \\
    V_5:& \quad f_5(x_1,x_2,x_3) = 30\frac{(x_1-1)(x_3-1)}{x_2^2(x_1-10)}, \notag \\
    & \quad x_{\text{train}} \sim U(0.05, 2, 300) \times U(1,2,300) \times U(0.05, 2, 300), \notag \\
    & \quad x_{\text{test}} \sim \tilde{E} \times E(0.95, 2.05, 0.1) \times \tilde{E}, \quad \tilde{E} = E(-0.05, 2.1, 0.15). \\
    V_6:& \quad f_6(x_1,x_2) = 6\sin x_1\cos x_2, \notag \\
    & \quad x_{\text{train}} \sim U(0.1, 5.9, 30)^2, \quad x_{\text{test}} \sim E(-0.05, 6.05, 0.02)^2. \\
    V_7:& \quad f_7(x_1,x_2) = (x_1-3)(x_2-3)+2\sin((x_1-4)(x_2-4)), \notag \\
    & \quad x_{\text{train}} \sim U(0.05, 6.05, 300)^2, \quad x_{\text{test}} \sim U(-0.25, 6.35, 1000)^2. \\
    V_8:& \quad f_8(x_1,x_2) = \frac{(x_1-3)^4+(x_2-3)^3-(x_2-3)}{(x_2-2)^4+10}, \notag \\
    & \quad x_{\text{train}} \sim U(0.05, 6.05, 50)^2, \quad x_{\text{test}} \sim E(-0.25, 6.35, 0.02)^2.
    \label{eq:VS8}
\end{align}

As in the original paper, the number of generations is set to 250 for all except two cases, $V_4$ and $V_6$, for which 500 are run. Every initial population is set to 100. As mentioned in Section \ref{sec:pysr}, all other hyperparameters are left as the PySR default. As in the original paper, three dictionaries are used:
\begin{enumerate}[label=(\alph*)]
    \item A basic function set with addition, subtraction, multiplication, division (all of which can freely operate on constants and variables) and squaring. 
    \item A set for the $V_1$ and $V_6$ problems which supplements the basic set with base-$e$ exponentiation. 
    \item A set for $V_2, V_3$ and $V_7$, as above but including $\sin$ and $\cos$.
\end{enumerate}
KAN \cite{liu2024kan20} is also tested on this benchmark. For the sake of consistency, it is run with a similar number of trainable parameters (after pruning) as the C-HiDeNN-TD in each case. For $V_1$, both are run with 40, $V_2$ 47 for C-HiDeNN-TD and 48 for KAN, $V_3$ 52 and 55, $V_4$ 75 and 110, $V_5$ 21 and 56, $V_6$ 10 and 40, $V_7$ 196 and 200, and $V_8$ C-HiDeNN-TD 20 and 40.
The results are shown in Table \ref{tab:rmse}. For each benchmark, we show the best-case median RMSE achieved in \cite{Vlad2009}, the median RMSE of five runs found by KAN, PySR and Ex-HiDeNN. All of these runs were performed on the host on an Apple M1 (8-core CPU, 8-core GPU, 8GB unified memory), and their mean time is reported. For $V_1, V_2, V_3, V_5$ and $V_6$, Ex-HiDeNN was able to find exact functional forms. This is perhaps unsurprising as their $\bar S_\otimes$ scores were all $1.00$. Note that the non-zero error for $V_6$ is due to its dictionary excluding trigonometric functions, but the exact expression is recovered with their reinclusion. Ex-HiDeNN sees particular savings in finding accurate expressions in the higher dimensional, near- or moderately separable functions $V_4, V_5$ whose $S_\otimes$ scores are $0.63$ and $1.00$, respectively. In fact, Ex-HiDeNN's median result was of the exact form for every multiplicative function. This is with the obvious exception of $V_6$, where the component functions were excluded from the dictionary. Across the board, KAN demonstrated an accuracy roughly on par with the genetic algorithm tested in \citet{Vlad2009}, but with extreme speed. However, in the tested configuration it did not compare in accuracy to PySR alone or Ex-HiDeNN.

\begin{table}[h]
    \centering
    \label{tab:rmse}
    \caption{Performance comparison of KAN, PySR, C-HiDeNN-TD and Ex-HiDeNN on benchmark data. KAN and C-HiDeNN-TD are initialized with the same number of trainable parameters. $^1$ denotes that the model recovered an exact functional form, up to rounding errors in constants. Results in bold denote the best performance on a given identifier.}
    \begin{tabular}{c|c|cc|cc|cc}
        \toprule
        Identifier & Ref & \multicolumn{2}{c|}{KAN} & \multicolumn{2}{c|}{PySR} & \multicolumn{2}{c}{Ex-HiDeNN} \\
        & & RMSE & Time/s & RMSE & Time/s & RMSE & Time/s \\
        \midrule
        $V_1$ & 0.069 & 0.244 & 6 & $0.018^1$ & 31 & $\textbf{0.000}^1$ & 44 \\
        $V_2$ & 0.212 & 0.598 & 5 & $0.042^1$ & 2 & $\textbf{0.000}^1$ & 54 \\ 
        $V_3$ & 0.739 & 1.005 & 10 & 0.078 & 175 & $\textbf{0.000}^1$ & 54 \\
        $V_4$ & 0.277 & 0.144 & 25 & $\textbf{0.000}^1$ & 568 & $0.006$ & 150 \\
        $V_5$ & 1.029 & 0.382 & 5 & $\textbf{0.000}^1$ & 52 & $\textbf{0.000}^1$ & 32 \\
        $V_6$ & 3.469 & 3.737 & 5 & 0.346 & 82 & $\textbf{0.163}$ & 68 \\
        $V_7$ & 1.457 & 2.056 & 17 & $1.365$ & 56 & $\textbf{0.081}^1$ & 69\\
        $V_8$ & 2.063 & 2.770 & 4 & 2.661 & 33 & $\textbf{0.532}$ & 23 \\
        \bottomrule
    \end{tabular}
\end{table}

\begin{figure}
    \centering
    \begin{tikzpicture}
        \node[anchor=south west,inner sep=0] (image) at (0,0) {\includegraphics[width=0.95\linewidth]{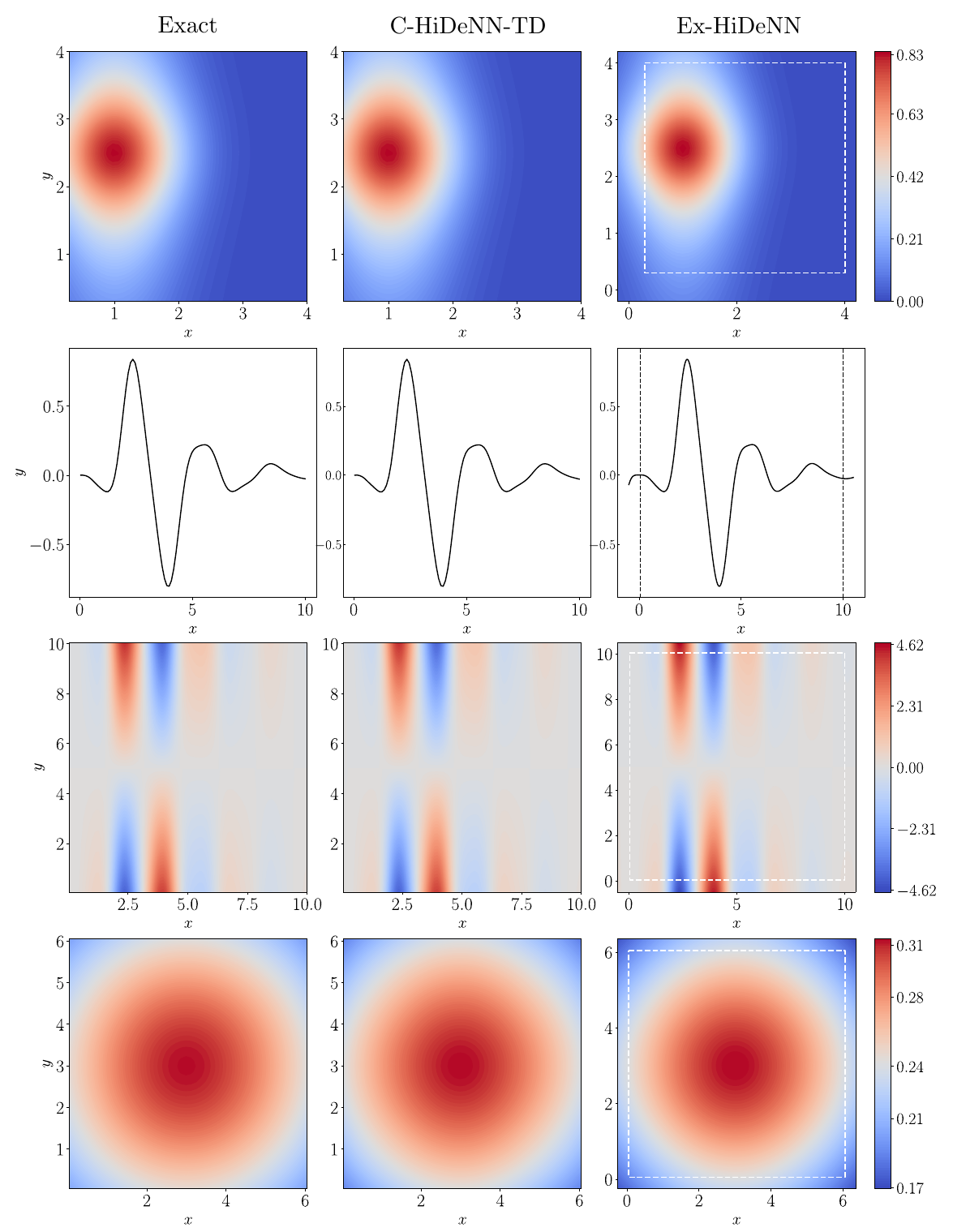}};
        \begin{scope}[shift={(image.south west)}, x={(image.south east)}, y={(image.north west)}]
            \node[anchor=east, font=\bfseries] at (0, 0.88) {$V_1$};
            \node[anchor=east, font=\bfseries] at (0, 0.635) {$V_2$};
            \node[anchor=east, font=\bfseries] at (0, 0.387) {$V_3$};
            \node[anchor=east, font=\bfseries] at (0, 0.135) {$V_4$};
        \end{scope}
    \end{tikzpicture}
    \caption{Exact, C-HiDeNN-TD, and Ex-HiDeNN results on the benchmark problems. From top to bottom, these represent the benchmarks $V_1, V_2, V_3$ and $V_4$.}
    \label{fig:fig5}
\end{figure}

\begin{figure}
    \centering
    \begin{tikzpicture}
        \node[anchor=south west,inner sep=0] (image) at (0,0) {\includegraphics[width=0.95\linewidth]{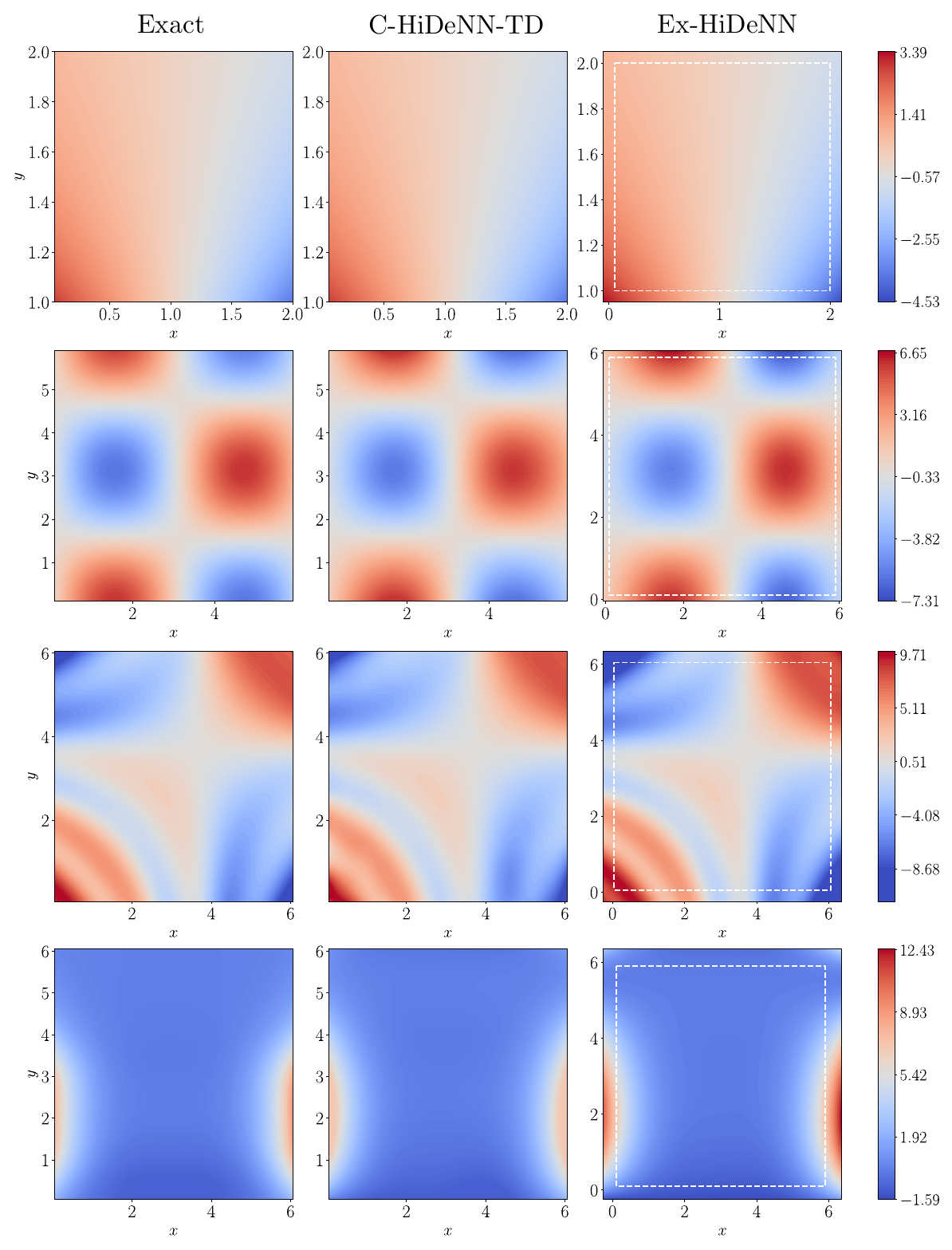}};
        \begin{scope}[shift={(image.south west)}, x={(image.south east)}, y={(image.north west)}]
            \node[anchor=east, font=\bfseries] at (0, 0.86) {$V_5$};
            \node[anchor=east, font=\bfseries] at (0, 0.62) {$V_6$};
            \node[anchor=east, font=\bfseries] at (0, 0.38) {$V_7$};
            \node[anchor=east, font=\bfseries] at (0, 0.139) {$V_8$};
        \end{scope}
    \end{tikzpicture}
    \caption{Exact, C-HiDeNN-TD, and Ex-HiDeNN results on the benchmarks problems. From top to bottom, these represent the benchmarks $V_5, V_6, V_7$, and $V_8$.}
    \label{fig:fig6}
\end{figure}
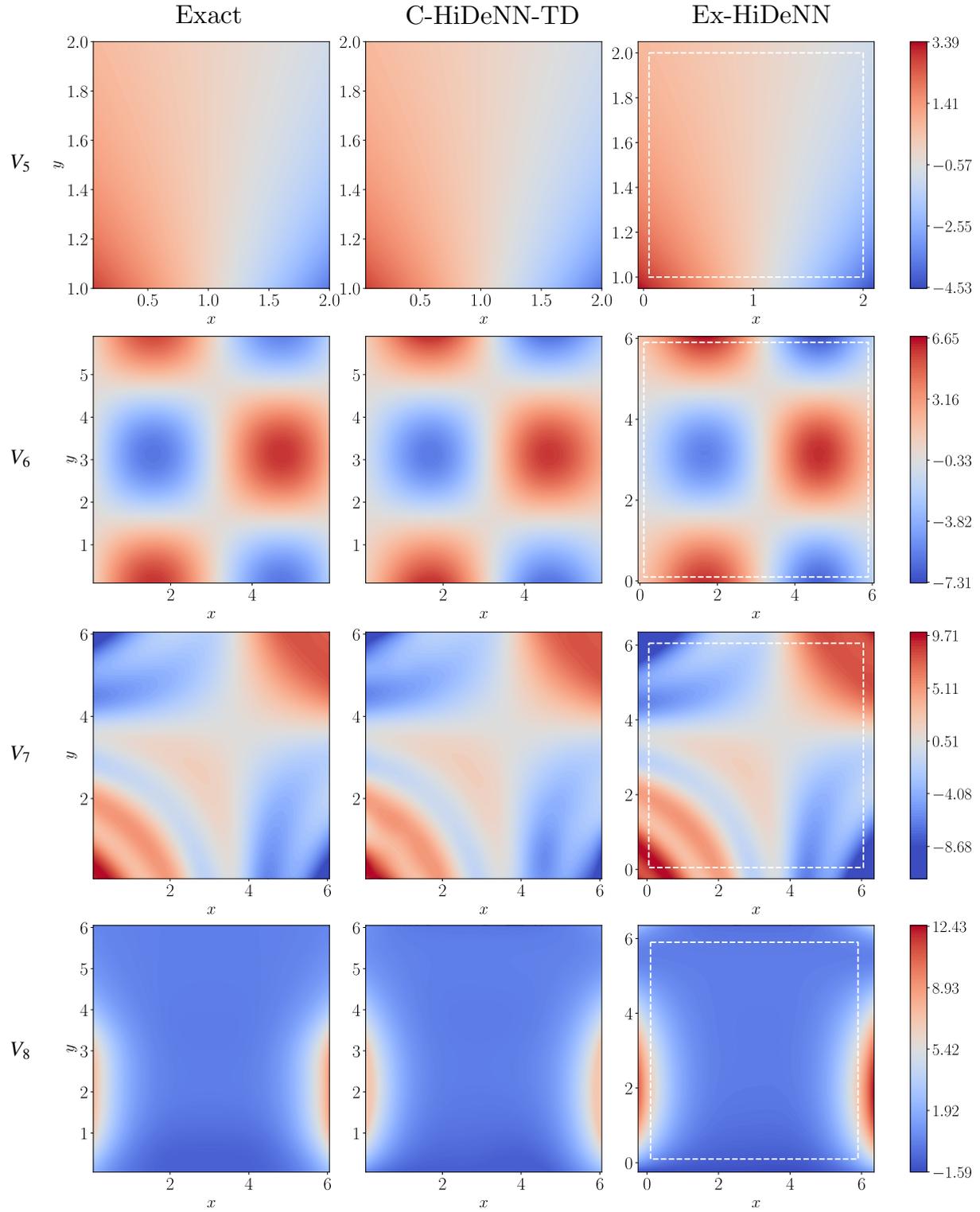

Figures \ref{fig:fig5} and \ref{fig:fig6} present plots for each function, showing the exact function, the C-HiDeNN-TD-derived surrogate, and the Ex-HiDeNN-derived form in its extrapolated test domain. The dashed border denotes the original training domain.

\clearpage

\clearpage
\subsection{Robustness to Noise: Comparison with Corrupted Benchmarks}

\begin{figure}
    \centering   \includegraphics[width=\linewidth]{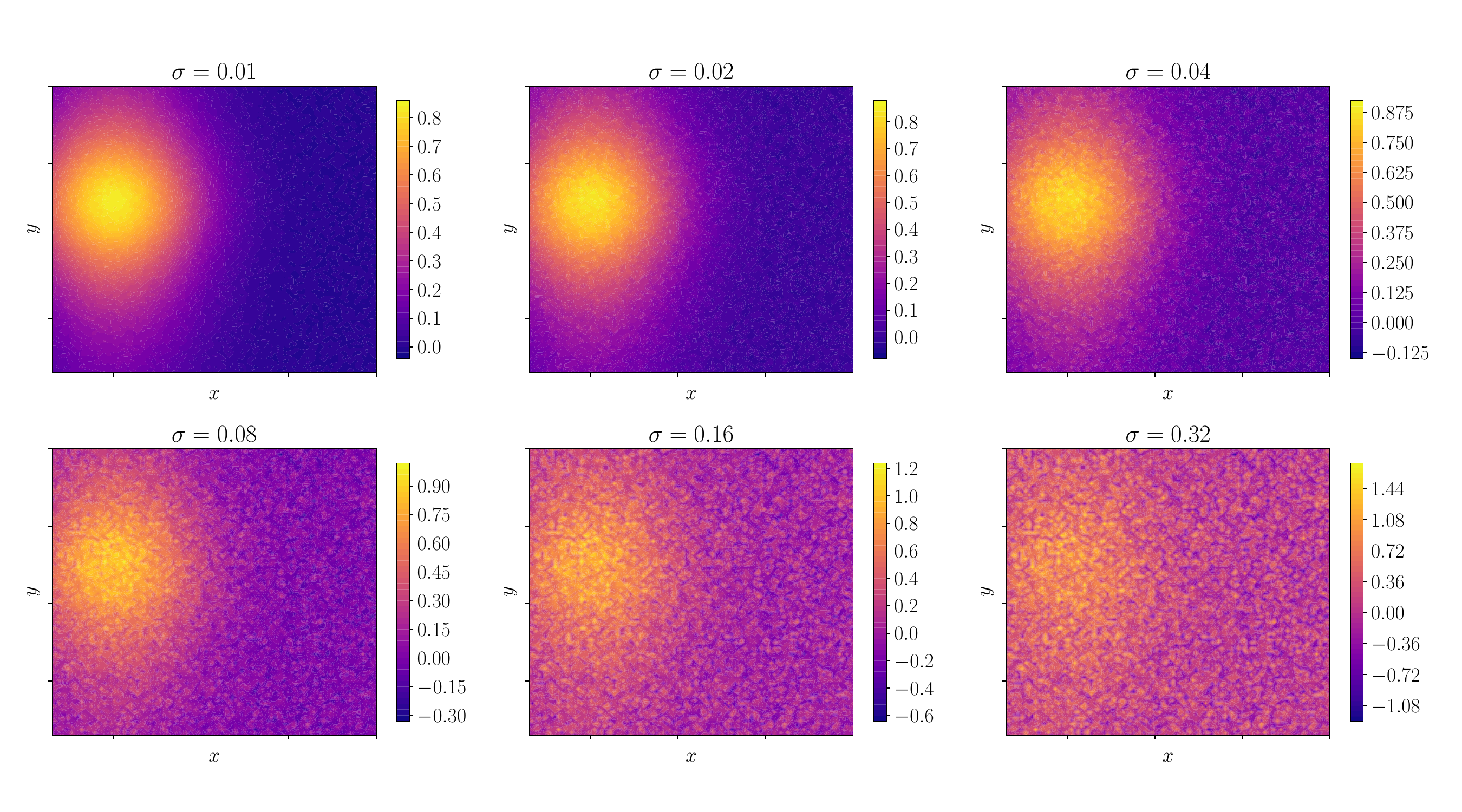}
    \caption{Plots visualizing the levels of corrupted data tested.}
    \label{fig:noisy}
\end{figure}

This sub-section evaluates the robustness of Ex-HiDeNN to noise in the dataset. Instead of re-doing all the benchmark examples, it is tested with the example $V_1$ on increasingly corrupted samples. However, it is trivial to extend to other benchmarks. We gradually added different levels of noise to the data and tested the efficacy of Ex-HiDeNN against PySR. The normalized samples have an added Gaussian noise term, $\varepsilon\sim\mathcal{N}(0,\sigma^2)$ with $\sigma=0.01, 0.02, 0.04, 0.08, 0.16$ and 0.32. These noisy data are visualized in Figure \ref{fig:noisy}. 

\begin{figure} [h]
    \centering    \includegraphics[width=0.8\linewidth]{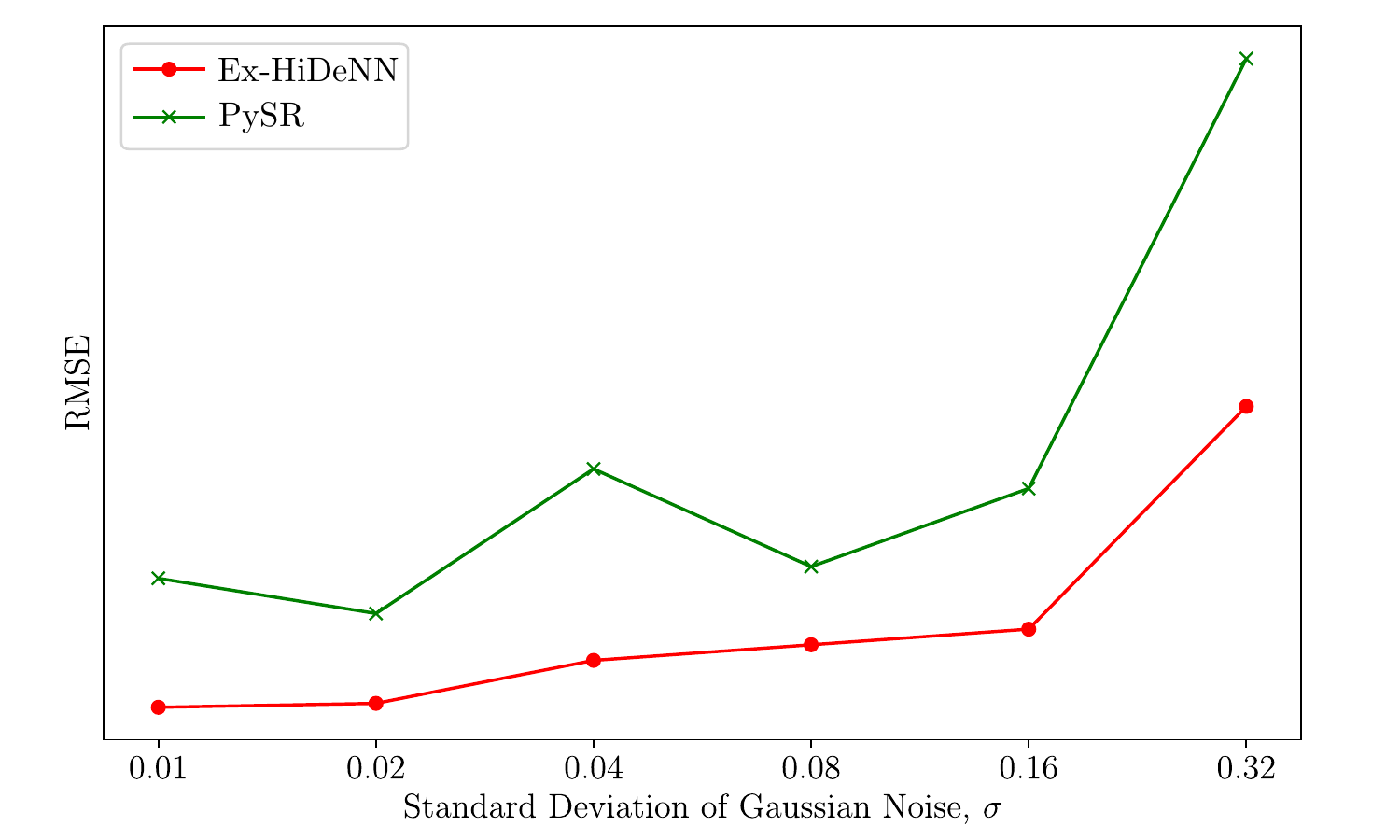}
    \caption{Plot showing the comparative performance of PySR alone and Ex-HiDeNN with increasing levels of corruption.}
    \label{fig:noisyresults}
\end{figure}

The result of the approximation performance for noisy data is shown in Figure \ref{fig:noisyresults}. A two-to-ten-fold improvement in RMSE is seen across the board. This is largely due to the interpolating nature of the surrogate: capturing the base signal while remaining robust to the noise. These results demonstrate Ex-HiDeNNs improved robustness to noise. 


\subsection{Discovery of Dynamical Systems with Ex-HiDeNN}
\label{sec:dyn}
\begin{figure}[h]
    \centering
    \label{fig:fig7}
   \includegraphics[width=0.55\textwidth]{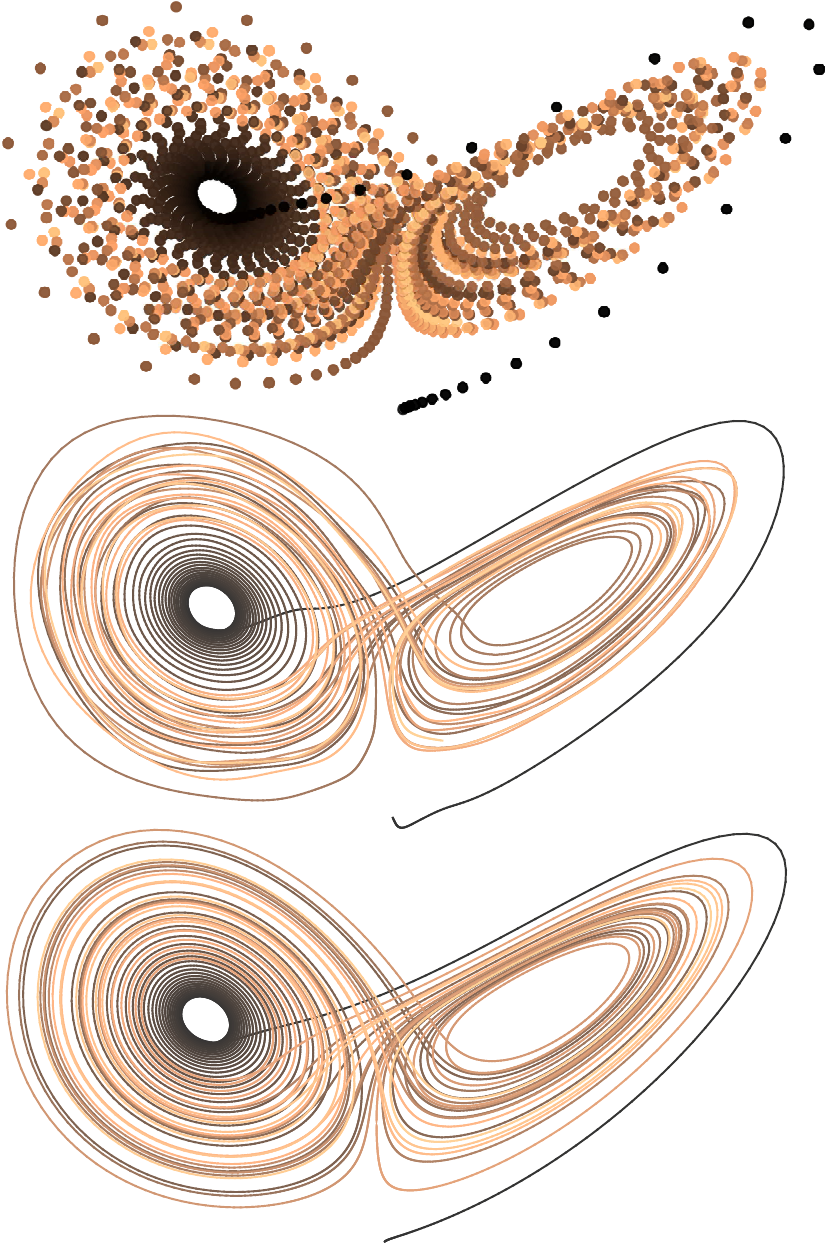}
    \caption{Process of discovery from a point cloud, to an automatically differentiable surrogate, to a simulation of the discovered dynamics. The trajectories start at black, and fade to orange as they progress.}
\end{figure}

In this part, we demonstrate how Ex-HiDeNN can be used to discover governing dynamics from time series data, somewhat akin to \cite{Brunton2016}, \cite{Naus2003}. We consider an $n-$dimensional dynamical system of the form,
\begin{align}
    \frac{d\boldsymbol{x}}{dt}=\boldsymbol{f}(\boldsymbol{x}(t)),
\end{align}
with $\boldsymbol{x}(t)\in\mathbb{R}^n$ a state vector/snapshot of the system at time $t$. Input data is in the form of snapshots over time. For times $t_1, t_2, \dots t_T$, this data is $\boldsymbol{x}(t_1),\boldsymbol{x}(t_2),\dots,\boldsymbol{x}(t_T)$ with approximated derivatives $\dot{\boldsymbol{x}}(t_1),\dot{\boldsymbol{x}}(t_2),\dots,\dot{\boldsymbol{x}}(t_T)$ Discovering governing dynamics is equivalent to finding a symbolic expression for $\boldsymbol{f}$, which is done by fitting derivative data to snapshot data. SINDy does this by constructing a "library" of nonlinear candidate function and finding coefficients by minimizing a regularized least squares term. Ex-HiDeNN constructs an interpolant from the data, $\boldsymbol{x}^h,$ and finds $\dot{\boldsymbol{x}}^h$ with automatic differentiation. Genetic programming then finds $\boldsymbol{f}$ by fitting $\dot{\boldsymbol{x}}^h$ to $\boldsymbol{x}^h$. 

As a demonstration, we take the Lorenz system. This is a well-known chaotic system described by three nonlinear differential equations,
\begin{align}
    \frac{dx}{dt}&=\sigma(y-x),\\
    \frac{dy}{dt}&=x(\rho-z)-y,\\
    \frac{dz}{dt}&=xy-\beta z,
\end{align}
with a typical choice of parameters being $\sigma=10, \beta = 8/3$ and $\rho=28$. Data is generated by numerically integrating from $t=0$ to $t=50$ with timestep $\frac{1}{40}$, taking $\boldsymbol{x}(0)=\left[\frac{1}{2},\frac{1}{2},\frac{1}{2}\right]^T$. With this, a $1-$input, $3-$output interpolating neural network is trained with 581 parameters per output. The interpolant is trained for 1708 epochs where it converges with a normalized RMSE of $0.009$. The interpolant is differentiated by automatic differentiation and PySR is run three times, once for each dimension, to fit it with each interpolant $\hat x, \hat y$ and $\hat z$. PySR is run with initial populations of 30, 100 and 100 for 100, 200 and 200 generations with a maximum complexity of 9, 11 and 9 in $x-,y-$ and $z$. The operator dictionary is fixed at $\{+,-,\times,\div,\cos,\sin,\exp\}$. 

Ex-HiDeNN recovers the exact equations to within two significant figures,
\begin{align}
    \frac{d\hat x}{dt}&=9.97(\hat y-\hat x),\\
    \frac{d\hat y}{dt}&=\hat x(27.9-\hat z)-\hat y,\\
    \frac{d\hat z}{dt}&=\hat x\hat y-2.67 \hat z.
\end{align}

\begin{figure}
    \centering
    \label{fig:fig10derivs}
     \includegraphics[width=0.85\textwidth]{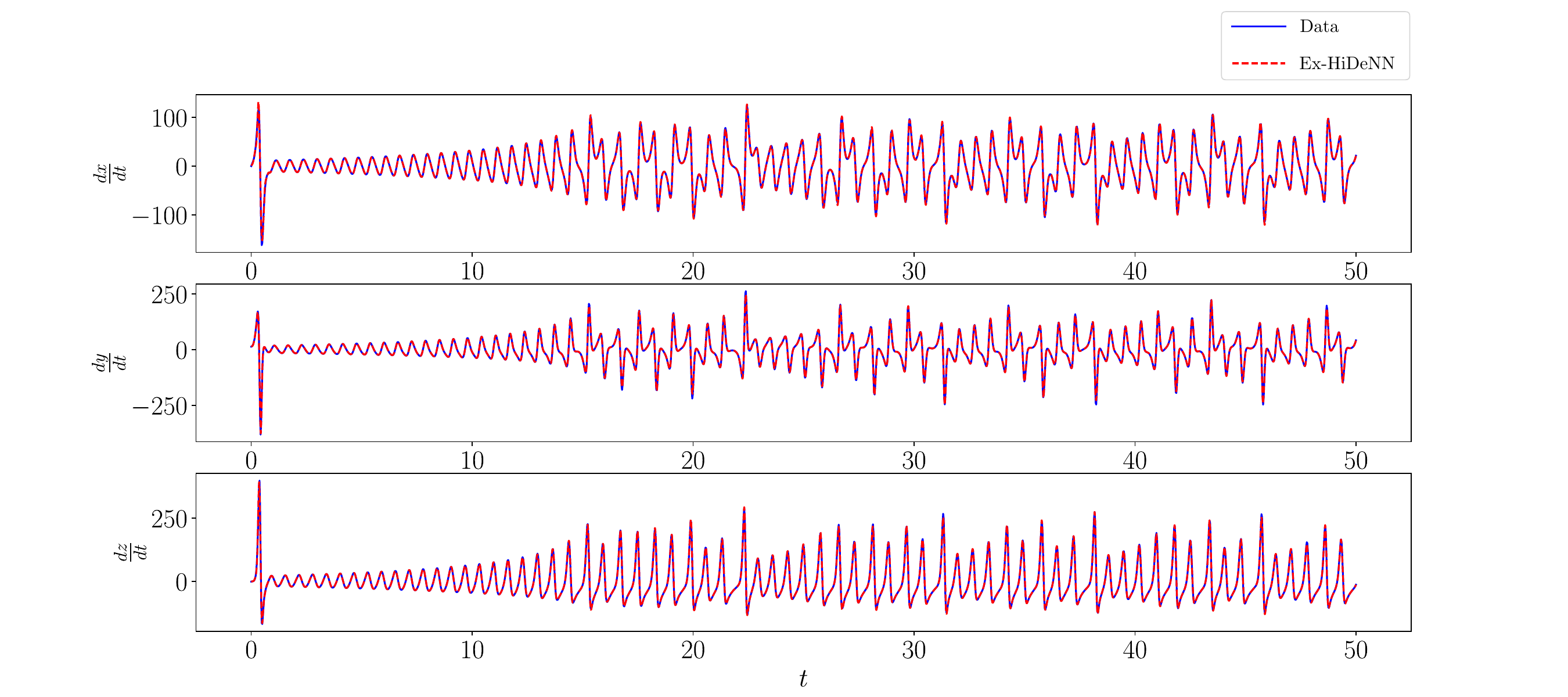}
    \caption{Overlay of the derivatives symbolically fitted by Ex-HiDeNN onto the exact data}
\end{figure}

The process from data, to a continuously differentiable interpolant, to simulated dynamics is shown in figure \ref{fig:fig7}. Figure \ref{fig:fig10derivs} overlays the learned forcing functions onto their exact counterparts, demonstrating an extremely tight fit.

\section{Applications in Current Engineering Problems}
\label{sec4}
To demonstrate Ex-HiDeNN's capacity for extracting interpretable and highly accurate models from experimental data, it is applied to three datasets taken from contemporary problems in engineering. In Section \ref{sec:fatigue} the NIMS dataset for rotation-bending fatigue of additively manufactured alloys, taken from \cite{Agrawal2014}, is considered. In Section \ref{sec:hardness}, the considered dataset from \cite{Tantardini2024} involves the prediction of Vickers Hardness from other material properties. Finally, Section \ref{sec:yield} investigates discovery of a constitutive law from the Matsuoka-Nakai dataset \cite{Bahmani2024}.
\clearpage
\subsection{Closed-Form Fatigue Equation for Additively Manufactured Alloys} 
\label{sec:fatigue}

Ex-HiDeNN's ability to model high-dimensional and sparse data is put to the test; the NIMS dataset takes 25 input parameters, as outlined in table \ref{tab:nims_features}, and gives experimental bending-rotating fatigue life in factors of $10^7$ cycles. The dataset includes 437 such experiments. This gives an approximate $437^{(\frac1 {25})}=1.27\dots$ samples per dimension - an extremely sparse and testing dataset.

\begin{table}[h]
\centering
\caption{NIMS Data Features for Fatigue Life Prediction}
\begin{tabular}{ll}
\hline
\textbf{Abbreviation} & \textbf{Details} \\
\hline
C      & \% Carbon \\
Si     & \% Silicon \\
Mn     & \% Manganese \\
P      & \% Phosphorus \\
S      & \% Sulphur \\
Ni     & \% Nickel \\
Cr     & \% Chromium \\
Cu     & \% Copper \\
Mo     & \% Molybdenum \\
NT     & Normalizing Temperature \\
THT    & Through Hardening Temperature \\
THt    & Through Hardening Time \\
THQCr  & Cooling Rate for Through Hardening \\
CT     & Carburization Temperature \\
Ct     & Carburization Time \\
DT     & Diffusion Temperature \\
Dt     & Diffusion Time \\
QmT    & Quenching Media Temperature (for Carburization) \\
TT     & Tempering Temperature \\
Tt     & Tempering Time \\
TCr    & Cooling Rate for Tempering \\
RedRatio & Reduction Ratio (Ingot to Bar) \\
dA     & Area Proportion of Inclusions Deformed by Plastic Work \\
dB     & Area Proportion of Inclusions Occurring in Discontinuous Array \\
dC     & Area Proportion of Isolated Inclusions \\
Fatigue & Rotating Bending Fatigue Strength ($10^7$ Cycles) \\
\hline
\end{tabular}
\label{tab:nims_features}
\end{table}

\begin{figure}[h]
    \centering
    \label{fig:fig11fatigue}
     \includegraphics[width=\textwidth]{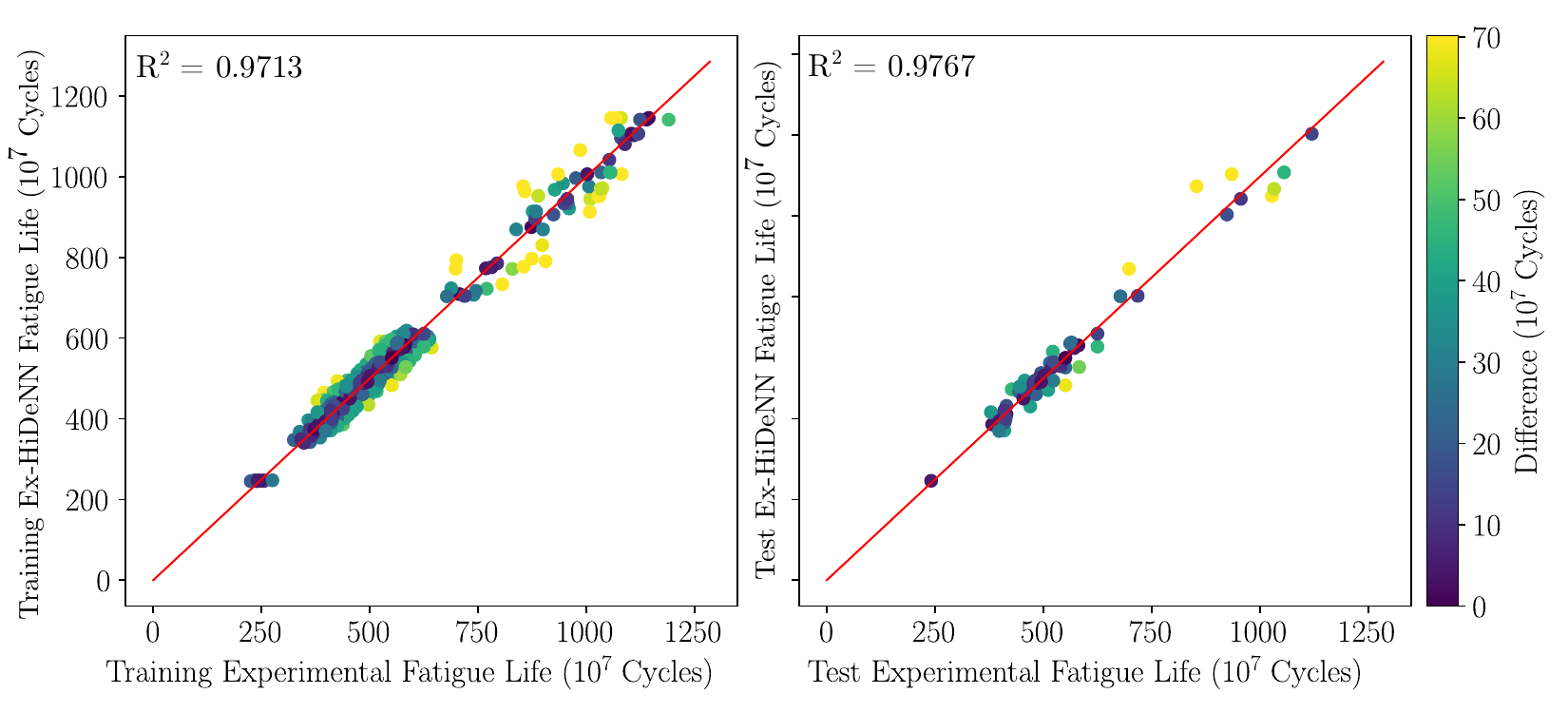}
    \caption{A closed-form equation for fatigue life of additively manufactured alloys. Correlation plots for the training set (\textit{left}) and test set (\textit{right}). These plots show the difference between the number of cycles found by experiment and those predicted by equation \eqref{eq:Fatigue}.}
\end{figure}

The data is split in an 80/10/10 train-validation-test split. An interpolating neural network with 600 trainable parameters (6 modes, 4 nodes per each of the 25 input dimensions) is trained for 2000 epochs. This surrogate returns a training RMSE of 19 and relative error of 1.9\%. The surrogate is then sampled at 520 points, with which we perform symbolic regression. The resulting expression is 
\begin{align}
\label{eq:Fatigue}
\textrm{Fatigue Life} = \left(\mathrm{CT} + 132 \log(\mathrm{C}) + 132 \sin(\mathrm{Cr}) \right) \exp\left(2.88\mathrm{dA}-\mathrm{Cu}\right) + 12300\frac{\mathrm{TCr} + \mathrm{Si} + \mathrm{Ni} + 0.75}{\mathrm{TT}}
\end{align}
Equation \eqref{eq:Fatigue} predicts the training set with a RMSE of 34, a relative error of 4.5\% and an $R^2$-score of 0.9713. In fact, this expression demonstrates impressive generalization to the test set, with an RMSE of 20, relative error of 2.8\% and an $R^2$-score of 0.9767. Correlation plots for each of the train and test sets are laid out in figure \ref{fig:fig11fatigue}. The discovery of a single, parsimonious expression that accurately models this complex, 25-dimensional, sparse data demonstrates the power of Ex-HiDeNN in finding underlying relationships in what is otherwise an intractable dataset.

In trying to understand the expression, it can be seen that carburization temperature (CT), cooling rate for tempering (TCr) and silicon and nickel contents (Si, Ni) all enter additively - all directly boosting fatigue life. Carbon and chromium contents (C, Cr) enter through concave functions (log, sin), so that their presence increases fatigue life with a diminishing effect. Additionally, the tempering temperature (TT) encodes a hyperbolic decay: low TT amplifies the alloying/TCr combination, while high TT quenches it. Meanwhile, the exponential multiplier splits into two pieces, $\exp(2.88$dA) - an exponential growth in dA - and $\exp($-Cu), an exponential decay with copper content. Together, these functional forms capture intuitive metallurgical realities: Cu softens steel in a saturating manner, while C, Cr, Si and Ni strengthen it. Tempering with rapid cooling and high carburization temperatures have life-extending effects.

\subsection{Hardness equation from Microindentation/Nanoindentation Data}
\label{sec:hardness}
\begin{figure}[h]
    \centering
    \label{fig:fig9}
     \includegraphics[width=\textwidth]{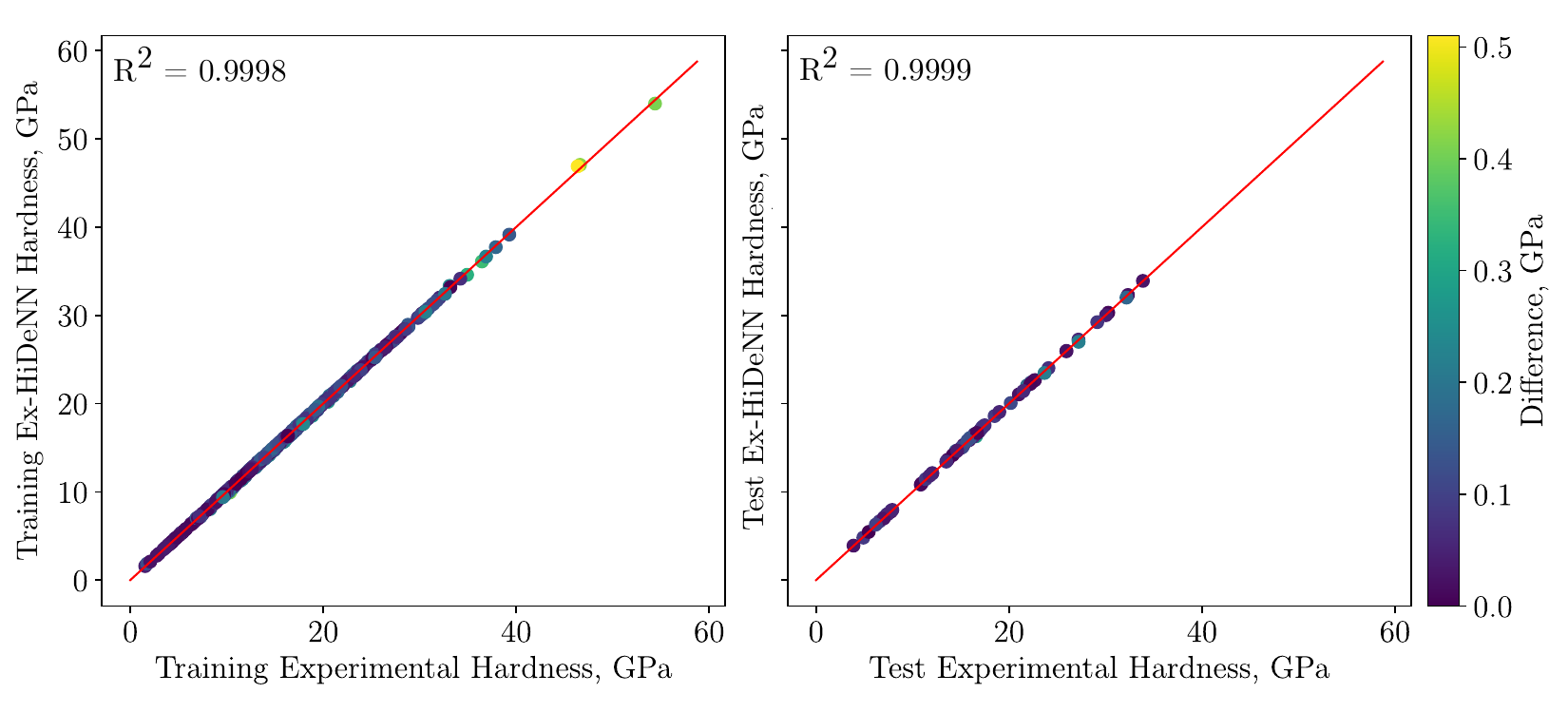}
    \caption{Correlation plots for the training set (\textit{left}) and test set (\textit{right}). These plots show the difference in Vickers Hardness found by experiment and those predicted by equation \eqref{e:HEx-HiDeNN}, in GPa. The diagonal, red line denotes perfect agreement.}
\end{figure}

To further validate Ex-HiDeNN on sparse data, this section focuses on work done by \cite{Tantardini2024}. Their dataset provides six main parameters for the prediction of Vickers Hardness, shown in table \ref{tab:microindentation}.
\begin{table}[h]
\centering
\caption{Parameters used in \cite{Tantardini2024} for the prediction of Vickers Hardness}
\begin{tabular}{ll}
\toprule
\textbf{Abbreviation} & \textbf{Details} \\
\midrule
$B_v$ & Voigt averaging of bulk modulus, GPa\\
$B_r$ & Reuss averaging of bulk modulus, GPa\\
$G_r$ & Reuss averaging of shear modulus, GPa\\
$\sigma$ & Poisson's ratio\\
$R_x$ & Maximum atomic radius, Å\\
$A_w$ & Weighted atomic weight, a.m.u\\
\bottomrule
\label{tab:microindentation}
\end{tabular}
\end{table}
A 630-parameter (15 modes, 7 nodes per each of the six feature dimensions) interpolating neural network is trained for 4000 epochs. The resulting surrogate has a normalized training RMSE of 0.015, and is sampled at 621 points using a combination of Lating Hypercube and jitter sampling.

\begin{align}
    \label{e:HEx-HiDeNN}
    H^{\textrm{Ex-HiDeNN}}
    =
    \frac{B_v}{\sigma}\left(\dfrac{0.396}{\log(G_r+1.08)+1.51}-0.03\right) - 
    0.146\frac{B_v R_x}{A_w}
- 5.61
\end{align}
When compared to their expression found by SISSO, 
\begin{align}
    H^\textrm{SISSO}=0.147\frac{B_V}{\sigma\sqrt[3]{G_R}}-1.136\frac{B_R\log R_X}{A_W}-5.679,
\end{align} some similar inclusions can be identified, $\frac{1}{A_W}, \frac{B_V}{\sigma}$, as well as a similarly saturating term in $G_r$ ($\log(G_r+1.08)$ compared to $\sqrt[3] {G_r}$). On the training set, \eqref{e:HEx-HiDeNN} returns a fitting RMSE of 100MPa, a maximum absolute error of 510MPa and a relative error of 0.51\%. The expression has an $R^2$-score of 0.9998. Figure \ref{fig:fig9} presents correlation plots of the Ex-HiDeNN-predicted Vickers Hardness against experimental values both on the training and test sets. Both plots show almost perfect alignment along the parity line, indicating that Ex-HiDeNN's discovered form generalizes nearly flawlessly from the training to unseen data. Specifically, the test set we find a RMSE of 90MPa, a relative error of 0.49\% and an $R^2$-score of 0.9999. This $R^2-$score indicates that the discovered expression explains essentially all of the variance in the held-out hardness data. Paired with the low RMSE, this near-perfect agreement goes beyond a well-fit model, but indicates low experimental scatter in the Vickers Hardness measurements.

To provide a direct comparison with SISSO, both models are evaluated on the full dataset. Ex-HiDeNN achieved a near-perfect overall prediction with an $R^2$-score of 0.9999 and RMSE of only 0.066GPa. In contrast, SISSO's learned expression yielded an $R^2$-score of 0.9552 and an RMSE of 1.66GPa. Ex-HiDeNN therefore demonstrated a 25-fold RMSE reduction, showcasing its ability to discover robust and highly accurate functional forms from data.

\begin{figure}[h]
    \centering
    \label{fig:fig10}
     \includegraphics[width=\textwidth]{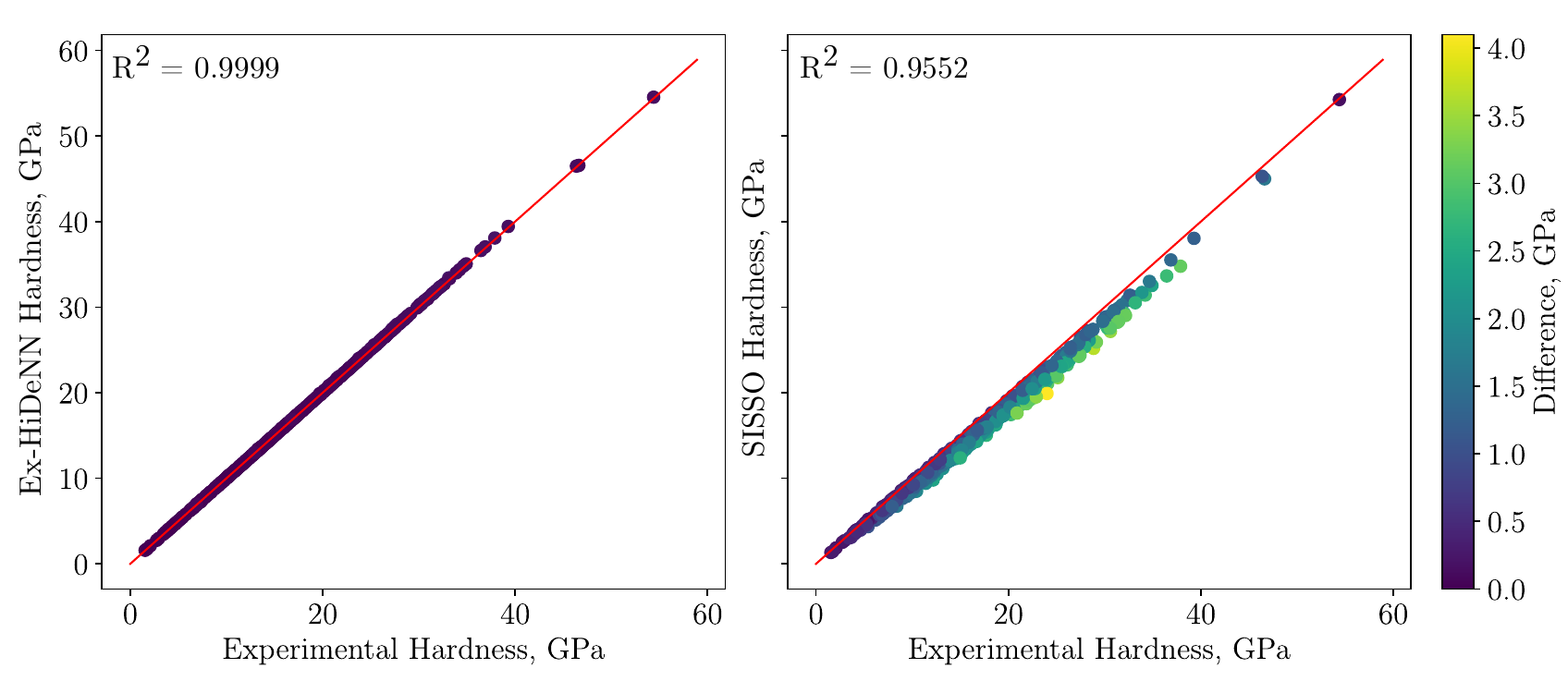}
    \caption{These are correlation plots for the entire dataset of hardness predicted by Ex-HiDeNN (\textit{left}) and by SISSO (\textit{right}). The diagonal, red line denotes perfect agreement.}
\end{figure}

Figure \ref{fig:fig10} presents correlation plots for the full microindentation dataset, comparing Ex-HiDeNN's and SISSO's models with the experimental Vickers hardness values. Ex-HiDeNN's results hug the perfect fit line - reflecting the near-zero bias and scatter across the entire hardness range. SISSO's predicions show visibly more spread, with errors reaching up to 4GPa. Together, these plots visually confirm the reported numerical metrics (RMSE, $R^2$).

\subsection{Yield Surface Expression from Matsuoka-Nakai Data}
\label{sec:yield}

Ex-HiDeNN is finally applied to discover classical constitutive laws, targeting the Matsuoka-Nakai yield surface. This is a three-invariant criterion used to model pressure-sensitive granular material yielding. Yield data, taken from \cite{Bahmani2024}, is generated by sampling at 20 equally spaced points in $p$ (pressure) between $[0,1000]$, 60 along the $\theta-$axis (lode angle) from 0 to $2\pi,$ and 11 from $0.85\rho$ to $1.15\rho$ (lode radius). Combinations of these parameters result in the three principal Cauchy stresses, $\sigma_1,\sigma_2$ and $\sigma_3$, the eigenvalues of the stress tensor so that $\sigma_1\geq\sigma_2\geq\sigma_3$. These then form the invariants, $I_1, I_2$ and $I_3$. This corresponds to 13,200, 9-dimensional input-output pairs, which are split 70/15/15 train/test/validation. 

\begin{figure}
    \centering
    \label{fig:fig12}
     \includegraphics[width=\textwidth]{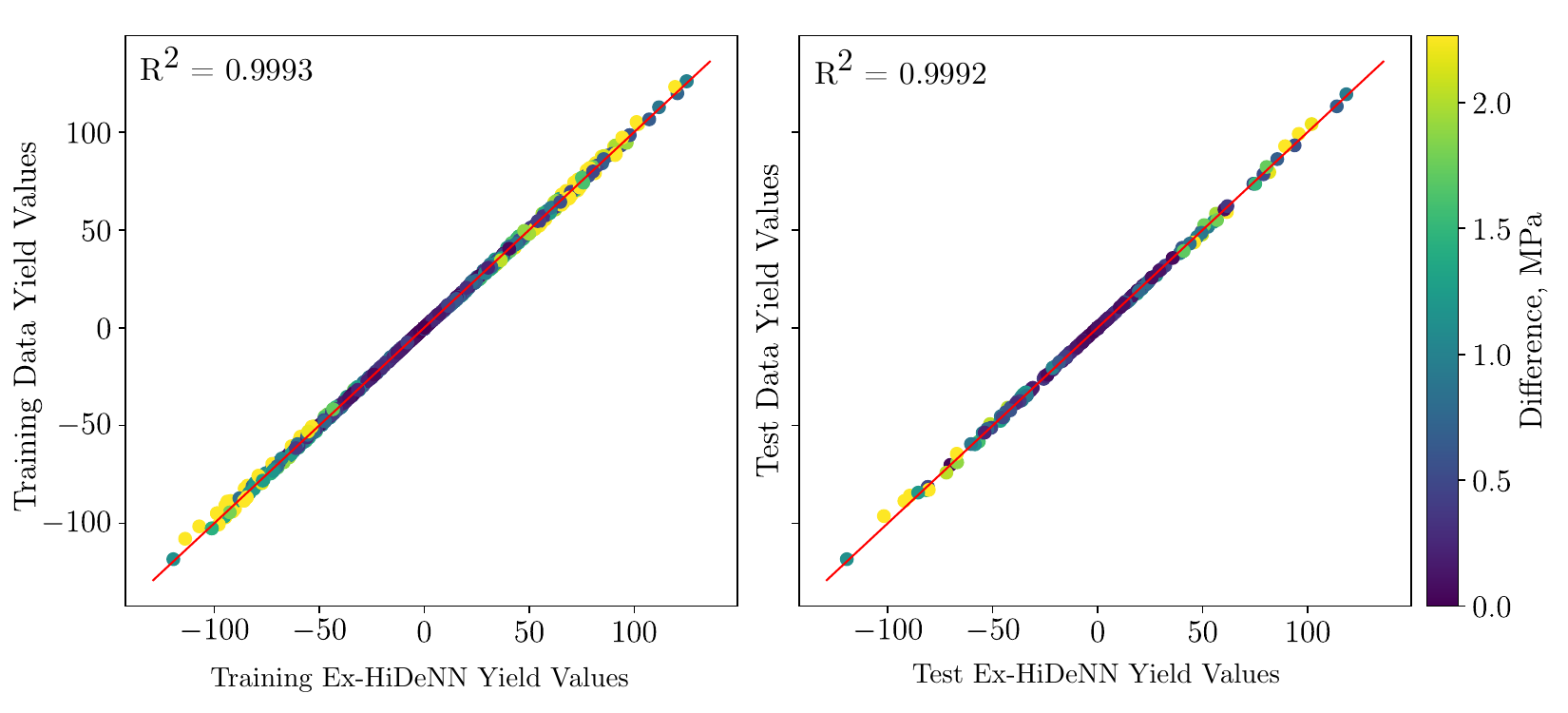}
    \caption{Correlation plots are shown for the training set (\textit{left}) and test set (\textit{right}). These plots show the difference in data yield values from those predicted by equation \eqref{e:HEx-HiDeNN}, in MPa. The diagonal, red line denotes perfect agreement.}
\end{figure}

Ex-HiDeNN trains on this data, learning the form
\begin{align}
    \label{eq:MN}
    f = 0.214\frac{\rho^2}{p}\log\left(1.251\sin\left(0.558 \frac{I_1 I_2}{I_3} \right) + 0.716\right),
\end{align}
with fitting RMSE 1.06, R$^2=0.9993$ and a relative error of 2.3\%. On the withheld test set, it achieved an RMSE of 1.10, an R$^2$-score of 0.9992 and a relative error of 2.3\%. Figure \ref{fig:fig12} shows parity plots for both the entire set and the training split, with each point colored by its absolute deviation from data in MPa. Again, we see near-perfect alignment down the parity line all the way across the yield range.

\begin{figure}[h]
    \centering
    \includegraphics[width=0.5\linewidth]{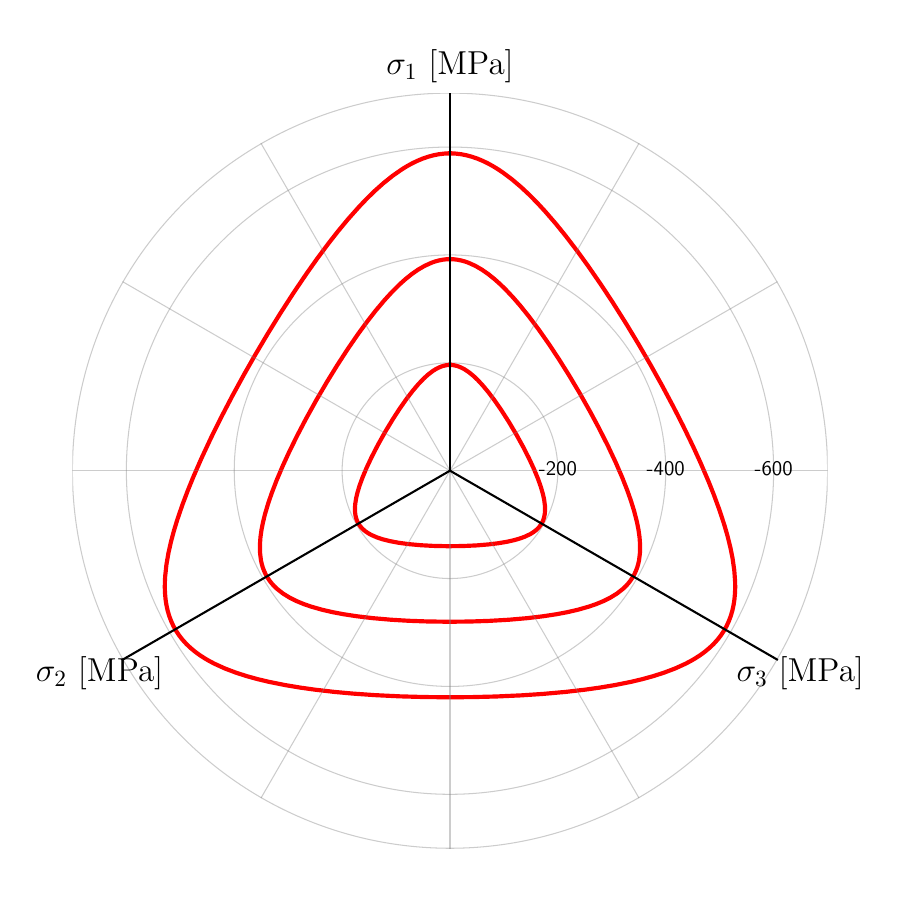}
    \caption{Deviatoric-plane yield loci at confining pressures $p=200,400$ and $600$MPa, obtained by the discovered expression \eqref{eq:MN}.}
    \label{fig:fig13}
\end{figure}

The expression itself demonstrates some satisfying features from a mechanics point of view. Firstly, the $\frac{\rho^2}{p}$ prefactor is a typical scaling seen in pressure-sensitive yield criteria. The logarithm term captures the diminishing effect seen in material strengthening. Crucially, the $\frac{I_1I_2}{I_3}$ argument, which is proportional to the lode angle, provides a clear rationale for the sine term as Matsuoka-Nakai surfaces are known to be lode-dependent. Further, on the yield surface at $f=0$, this form is reduced to $\frac{I_1I_2}{I_3}=\eta$, the classic Matsuoka-Nakai invariant condition. These yield surfaces for confining pressures of $p=200,400$ and $600$MPa are shown in Figure \ref{fig:fig13}

The ability of Ex-HiDeNN to derive a classical constitutive law directly from stress-strain data demonstrates its potential for accelerating discovery of new material model.

\section{Limitations and Future Work}
\paragraph{Dependence on C-HiDeNN-TD Surrogates}
The accuracy of the final expression strongly depends on the quality of the first-stage surrogate learned by C-HiDeNN-TD. If the C-HiDeNN-TD fails to capture the underlying function accurately, due to insufficient data, inappropriate C-HiDeNN-TD structure, or otherwise, the symbolic regression stage will produce poor quality expressions.
\paragraph{Challenges in Symbolic Regression}
It is important to note that even in one-dimensional cases, symbolic regression is an NP-hard problem - albeit significantly more tractable than in multidimensional cases. As with any algorithm, the success of PySR depends on the pre-defined dictionary of mathematical operators. If underlying functions call for operators not included in the dictionary, Ex-HiDeNN will struggle to discover them - see example $V_6$ in Section \ref{sec:bench}. Additionally, PySR can get stuck in local minima. It is not uncommon to see PySR alternate between two or even more forms on different runs.
\paragraph{Weaknesses of Ex-HiDeNN}
While Ex-HiDeNN has a built-in strategy for tackling low-separability data, its strength comes in exploiting separability. Scalability to high dimensions in low separability cases remains a hurdle, but the strategy of intelligent sampling from the surrogate provides savings over symbolic regression alone. Additionally, Ex-HiDeNN may struggle with discovering accurate forms from data with discontinuities or kinks.
\paragraph{Future Directions for Surrogate Modeling}
Given the strong dependence of Ex-HiDeNN's performance on the initial surrogate, a key avenue of immediate future work is the exploration of alternative surrogate models. Namely, there is a plan to integrate KHRONOS \cite{batley2025khronos}, a kernel-based neural architecture, as the first-stage surrogate in a new framework. KHRONOS's design may lead to more accurate and efficiently generated surrogates, particularly for complex, high-dimensional and strongly nonlinear data. Additionally, KHRONOS demonstrates significant robustness to noise.  Overall, the implementation of KHRONOS appears to be a promising avenue for exploration.
\paragraph{Future Directions for Symbolic Regression}
Another future direction is in the implementation of alternative state-of-the-art symbolic regression engines. A more fundamental and elegant evolution could entail the integration of a symbolic regression engine able to learn from the continuously differentiable surrogate provided by the C-HiDeNN-TD (or KHRONOS) stage, rather than relying on discrete sampling. A form of gradient-based expression discovery could be applied here, exploiting the automatically differentiable and structured nature of C-HiDeNN-TD or KHRONOS surrogates.

\label{sec5}
\section{Conclusion}

This paper introduced a novel, hybrid AI framework: Explainable Interpolating Neural Networks (Ex-HiDeNN). This framework was designed to address the critical need for interpretable models in complex engineering systems. Traditional data-driven approaches often result in ``black-box" models, while existing symbolic regression techniques face challenges in consistency and scalability, especially in high-dimensional problems. Ex-HiDeNN overcomes these limitations by combining the interpretability of symbolic regression with the expressive power of a structured surrogate model provided by an interpolating neural network (C-HiDeNN-TD). A key C-HiDeNN-TDovation, the use of the C-HiDeNN-TD to determine a pointwise separability score $S_\otimes$, guides the sampling and regression strategy in a way tailored to the inherent structure of the data.

Section \ref{sec2} laid out the interpolating neural network surrogate, the symbolic regression engine (PySR), and details of the Hessian separability score $S_\otimes$. Section \ref{sec3} validated Ex-HiDeNN on some contemporary benchmarks, demonstrating its ability to accurately and efficiently recover expressions from data, as well as its ability to recover a dynamical system (the Lorenz system) from data. Section \ref{sec4} carried out an investigation into Ex-HiDeNN's use in current problems in engineering. Section \ref{sec:fatigue} used it to uncover an expression for fatigue life in additively manufactured steel alloys with a relative error of $2.8\%$ on unseen test data. Section \ref{sec:hardness} had Ex-HiDeNN discover an expression predicting Vickers hardness from microindentation data. This expression had an exceptional relative error of $0.49\%$ on unseen test data. Finally, it was applied to learn a classical constitutive law for a Matsuoka-Nakai yield surface, used to model pressure-sensitive metal yielding. The form returned a relative error of $2.3\%$ on withheld test data.

The ability of Ex-HiDeNN to learn parsimonious, symbolic forms that are human-understandable, and accurate enough to be actionable represents a step towards more trustworthy AI in science and engineering applications. Such explicit mathematical forms allow for direct, physical interpretation and facilitate easy integration with existing theoretical frameworks and simulation pipelines. While the current implementation is not free from limitations, future work will focus on integrating advanced surrogates like KHRONOS, and tailoring the symbolic regression engine to better exploit surrogate properties.

\section{Acknowledgments}
\label{sec6}
S. Saha gratefully acknowledges the start-up fund provided by the by the Kevin T. Crofton Department of Aerospace and Ocean Engineering, Virginia Polytechnic Institute and State University. R. Batley acknowledges the Crofton Fellowship from the Kevin T. Crofton Department of Aerospace and Ocean Engineering, Virginia Polytechnic Institute and State University.
\clearpage

\bibliographystyle{elsarticle-num-names} 
\bibliography{references.bib}

\end{document}